\documentclass{article}

    \PassOptionsToPackage{numbers, compress}{natbib}

\usepackage[preprint]{neurips_2023}

\usepackage[utf8]{inputenc} 
\usepackage[T1]{fontenc}    
\usepackage{hyperref}       
\usepackage{url}            
\usepackage{booktabs}       
\usepackage{amsfonts}       
\usepackage{nicefrac}       
\usepackage{microtype}      
\usepackage{xcolor}         

\usepackage{graphicx}
\usepackage{subfigure}
\usepackage{multirow}
\usepackage{wrapfig}
\graphicspath{{figure/}}
\usepackage{algorithm}
\usepackage{algorithmic}
\usepackage{listings}
\title{G-Adapter: Towards Structure-Aware Parameter-Efficient Transfer Learning for Graph Transformer Networks}

\author{
    Anchun Gui, Jinqiang Ye and Han Xiao\thanks{Corresponding author.} \\
    Department of Artificial Intelligence \\
    School of Informatics, Xiamen University \\
    \texttt{\{anchungui,jinqiangye\}@stu.xmu.edu.cn, bookman@xmu.edu.cn} \\
}

\begin{document}

\maketitle

\begin{abstract}
It has become a popular paradigm to transfer the knowledge of large-scale pre-trained models to various downstream tasks via fine-tuning the entire model parameters. However, with the growth of model scale and the rising number of downstream tasks, this paradigm inevitably meets the challenges in terms of computation consumption and memory footprint issues. Recently, Parameter-Efficient Fine-Tuning (PEFT) (e.g., Adapter, LoRA, BitFit) shows a promising paradigm to alleviate these concerns by updating only a portion of parameters. Despite these PEFTs having demonstrated satisfactory performance in natural language processing, it remains under-explored for the question of whether these techniques could be transferred to graph-based tasks with Graph Transformer Networks (GTNs). Therefore, in this paper, we fill this gap by providing extensive benchmarks with traditional PEFTs on a range of graph-based downstream tasks. Our empirical study shows that it is sub-optimal to directly transfer existing PEFTs to graph-based tasks due to the issue of \textbf{\textit{feature distribution shift}}. To address this issue, we propose a novel structure-aware PEFT approach, named G-Adapter, which leverages graph convolution operation to introduce graph structure (e.g., graph adjacent matrix) as an inductive bias to guide the updating process. Besides, we propose Bregman proximal point optimization to further alleviate feature distribution shift by preventing the model from aggressive update. Extensive experiments demonstrate that G-Adapter obtains the state-of-the-art performance compared to the counterparts on nine graph benchmark datasets based on two pre-trained GTNs, and delivers tremendous memory footprint efficiency compared to the conventional paradigm.
\end{abstract}

\section{Introduction}
Pre-training then fine-tuning has become a prevalent training paradigm with the remarkable success of large-scale pre-trained models in Natural Language Processing (NLP) \cite{brown2020language, devlin2019bert, lewis2020bart, liu2019roberta, raffel2020exploring, touvron2023llama}. Recently, more researchers are striving to apply this paradigm to graph-based tasks with Graph Transformer Networks (GTNs) \cite{bo2023specformer, chen2022structureaware, chen2023nagphormer, diao2023relational, jin2023edgeformers, kim2022pure, kreuzer2022rethinking, maziarka2020molecule, mialon2021graphit, ying2021transformers, yuan2023monocular, zhang2020graphbert, zhao2021gophormer}. For instance, based on multi-layer Transformer encoders \cite{vaswani2017attention}, Graphormer \cite{ying2021transformers} first performs the well-designed unsupervised tasks on large-scale molecular datasets, and then fine-tunes the entire pre-trained parameters of the model on downstream molecular tasks of interest, which is also known as \textit{full fine-tuning}. However, full fine-tuning poses several issues in practice:
(\romannumeral 1) Given that the labels of graph data from some domains (e.g., chemistry, biology) are inaccessible without the expertise and labor-heavy annotations \cite{xia2022survey}, it is common that there are insufficient labeled samples in downstream tasks of interest. Hence, full fine-tuning would incur serious over-fitting and catastrophic forgetting issues \cite{houlsby2019parameterefficient, wang2021kadapter}.
(\romannumeral 2) When handling multiple diverse downstream tasks, full fine-tuning has to duplicate a modified copy of all parameters per task, which hinders the flexibility and applicability of large-scale models, especially in scenarios with constrained storage resources (e.g., mobile detection devices).

\begin{figure}[t]
    \centering
    \includegraphics[width=0.9\textwidth]{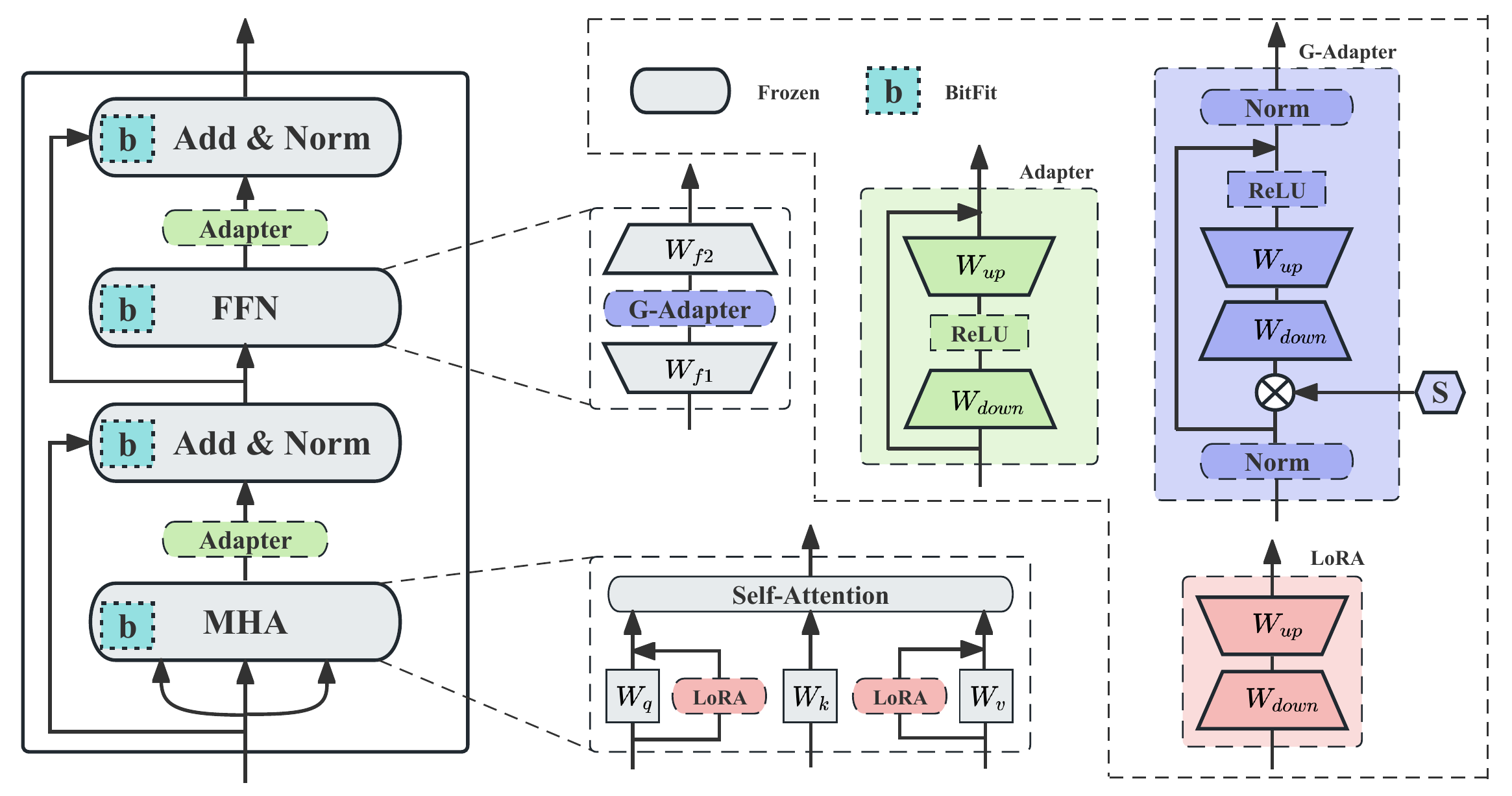}
    \caption{An overview of existing popular PEFTs (Adapter, LoRA, and BitFit) and our proposed G-Adapter. In the left part, we demonstrate the insertion position of PEFT blocks in a standard encoder of Transformer. In the right part, we depict the architecture of each PEFT method. Here, each color represents an approach, where the grey blocks are frozen during the fine-tuning process. Our proposed G-Adapter is marked and demonstrated in purple, in which \textbf{S} indicates the introduced graph structure information (e.g., the graph adjacent matrix).}
    \label{fig:overview}
\end{figure}

Recently, Parameter-Efficient Fine-Tuning (PEFT), as an alternative to full fine-tuning, has been proposed and widely investigated in NLP \cite{ben-zaken2022bitfit, ding2022delta, he2022unified, houlsby2019parameterefficient, hu2022lora}. PEFT aims to achieve competitive performance with full fine-tuning while consuming computation and storage resources as few as possible. Instead of updating the entire parameters during the fine-tuning phase, PEFT only updates a small fraction of parameters within the original model or additionally introduced modules, while freezing the remaining parameters. For example, Adapter \cite{houlsby2019parameterefficient} inserts two compact modules in each encoder of Transformer, while BitFit \cite{ben-zaken2022bitfit} only updates the bias terms in the model parameters, as shown in Fig.~\ref{fig:overview}. Despite the remarkable achievements of traditional PEFTs in natural language understanding tasks, the question is still under-explored whether these PEFTs from the language domain are feasible for various GTNs under graph-based tasks, given that the intrinsic discrepancy between graph and text modalities (e.g., the graph has rich structure information). Therefore, in this paper, we shall fill this gap by answering the following questions: \textbf{Can PEFTs from the language domain be transferred directly to graph-based tasks? If not, how to design a graph-specific PEFT method?}

\begin{figure}[t]
    \centering
    \subfigure[On large-scale datasets.]{
        \includegraphics[width=0.45\textwidth]{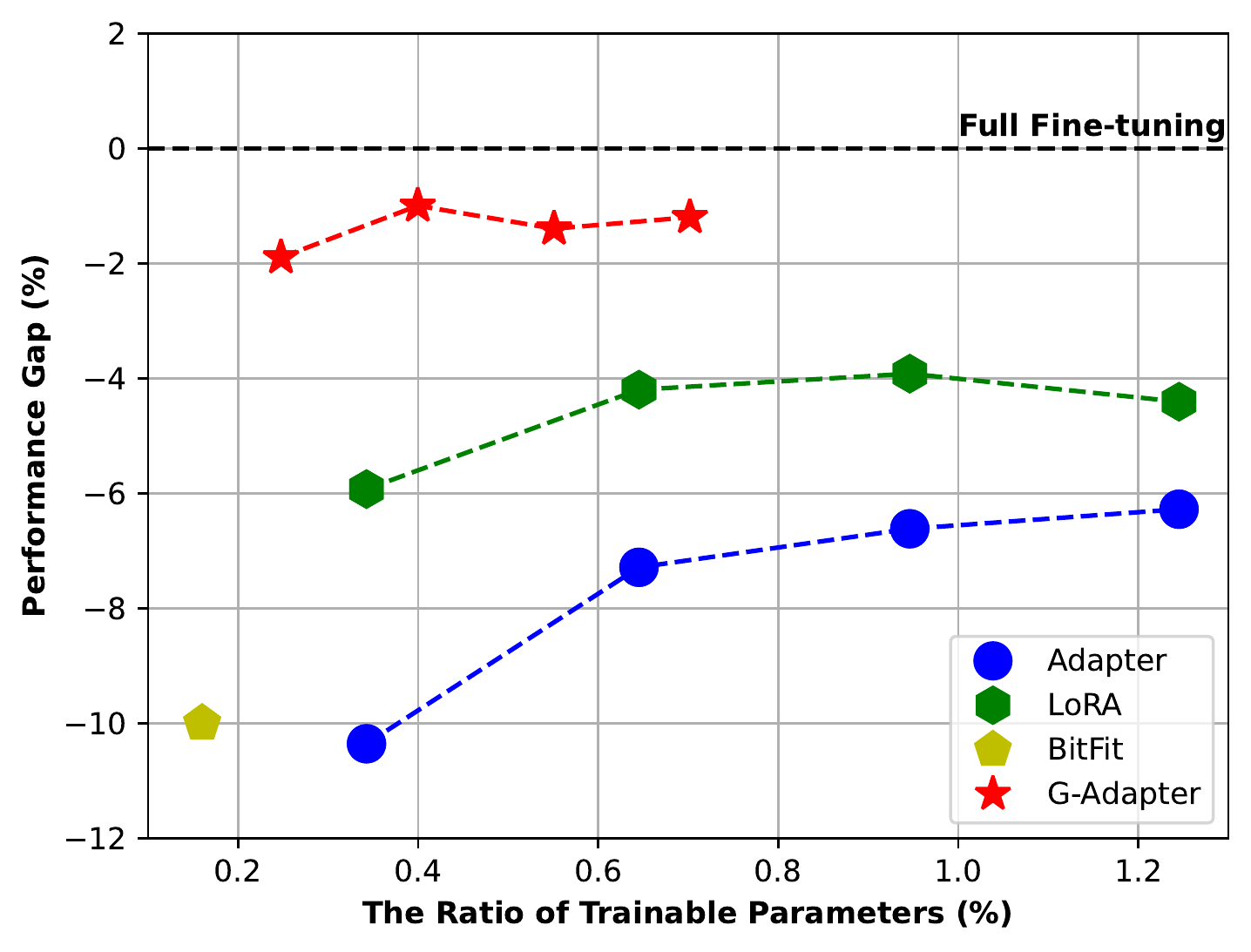}
    }
    \subfigure[On small-scale datasets.]{
        \includegraphics[width=0.45\textwidth]{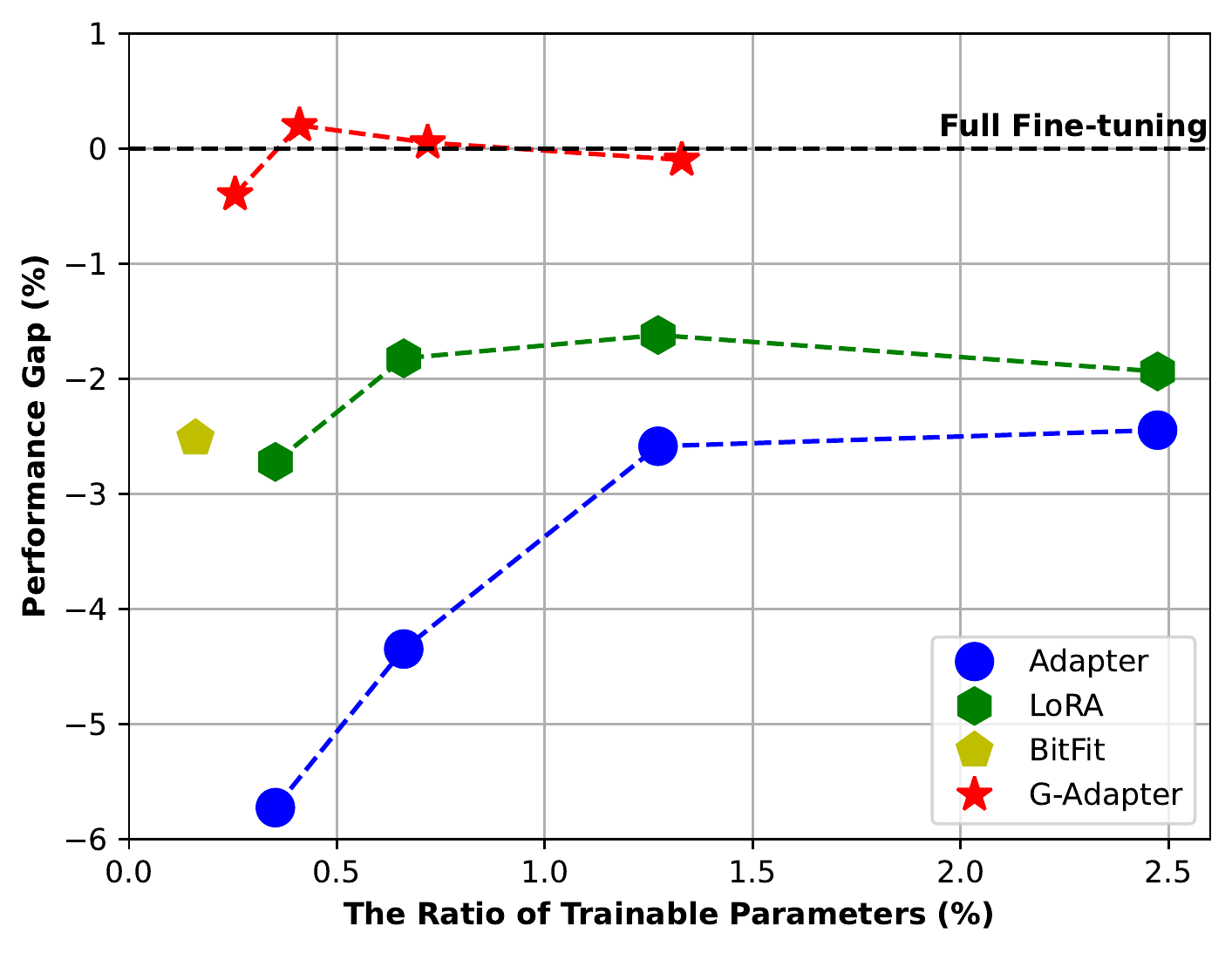}
    }
    \caption{The comparison between PEFTs (Adapter, LoRA, BitFit, and G-Adapter) and full fine-tuning on large- and small-scale datasets. (a) Based on the pre-trained Graphormer, we first average the results of each PEFT on two large-scale datasets, and then compute the performance gap compared to full fine-tuning. (b) Similarly, we calculate the performance gap of each PEFT on seven small-scale datasets, based on another pre-trained model MAT. Refer to Sec.~\ref{subsec:pilot_exp} for more descriptions.}
    \label{fig:intro_compare}
\end{figure}

\begin{figure}[t]
    \centering
    \includegraphics[width=1\textwidth]{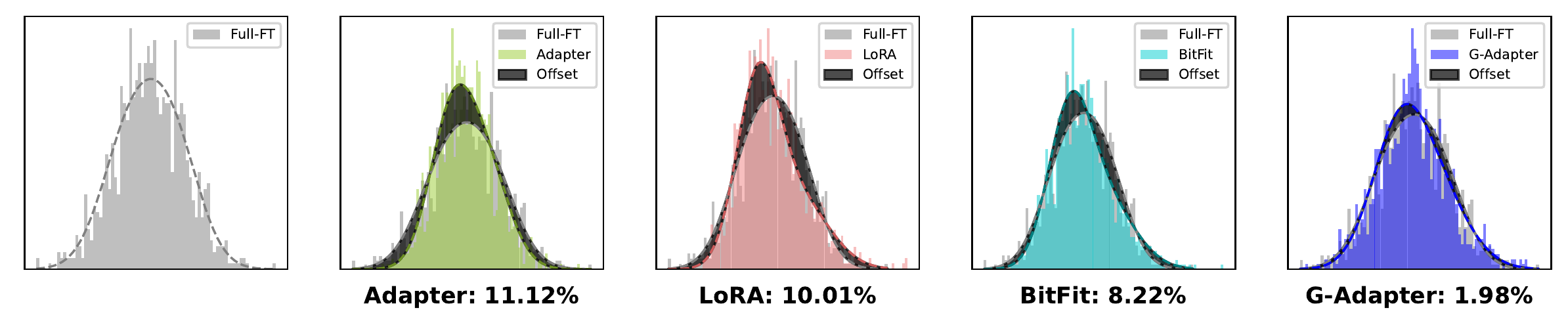}
    \caption{The illustration of feature distribution shift, where Full-FT denotes full fine-tuning. For the identical input, the feature distribution of traditional PEFTs (Adapter, LoRA, and BitFit) has a significant offset (dark region) compared to full fine-tuning. In contrast, our proposed G-Adapter has a highly similar behavior to full fine-tuning. Here, Jensen-Shannon divergence is utilized to measure the discrepancy between two distributions. Refer to Sec.~\ref{subsec:pilot_exp} for more discussions.}
    \label{fig:feat_dist_shift}
\end{figure}

To start with, we comprehensively examine the performance of mainstream PEFTs (Adapter \cite{houlsby2019parameterefficient}, LoRA \cite{hu2022lora}, and BitFit \cite{ben-zaken2022bitfit}) on popular molecular graph datasets based on two pre-trained GTNs (Graphormer \cite{ying2021transformers} and MAT \cite{maziarka2020molecule}). The overall comparison is shown in Fig.~\ref{fig:intro_compare}, in which we unfortunately observe a significant gap between traditional PEFTs and full fine-tuning, especially on large-scale datasets. Further, our exploration reveals the \textit{feature distribution shift} issue due to the absence of graph structure in the fine-tuning process (see Fig.~\ref{fig:feat_dist_shift} and Sec.~\ref{subsec:pilot_exp} for more discussions). To alleviate these concerns, we propose a novel structure-aware PEFT method, G-Adapter, which leverages graph convolution operation to introduce graph structure as the inductive bias to guide the updating process. Moreover, we apply the low-rank decomposition to the learnable weights, which makes G-Adapter highly lightweight. Besides, we propose Bregman proximal point optimization to further ease the feature distribution shift by preventing the model from aggressive update.

To verify the effectiveness of our approach, we conduct extensive experiments on a variety of graph-based downstream tasks based on pre-trained GTNs. The results demonstrate that our proposed G-Adapter can effectively address the feature distribution shift issue and significantly enhance the performance. Specifically, 
(\romannumeral 1) G-Adapter obtains the state-of-the-art performance than baselines on both large- and small-scale datasets. Even compared to full fine-tuning, G-Adapter could achieve competitive (or superior) results with fewer trainable parameters. For example, full fine-tuning achieves $0.804$ AUC with $100\%$ trainable parameters on MolHIV, while G-Adapter gains $0.790$ AUC with only $0.24\%$ trainable parameters.
(\romannumeral 2) G-Adapter enjoys remarkable advantages over full fine-tuning in terms of memory footprint. For instance, full fine-tuning stores $161$MB checkpoint per task while G-Adapter merely requests $0.4$MB\footnote{Here, the bottleneck size $r = 4$, based on the pre-trained MAT. See Tab.~\ref{tab:eff_inf_mem} for more comparisons.} checkpoint for each downstream task. Additionally, the introduced G-Adapter modules barely degrade the training efficiency and inference speed, and extensive ablation experiments also confirm the rationality of each component in our design.

To summarize, our contributions are as follows:
\begin{itemize}
    \item To the best of our knowledge, this is the first work formally to investigate the parameter-efficient fine-tuning of graph-based tasks and models. And, we benchmark several widely used PEFTs from the language domain on a range of graph-based downstream tasks.
    \item We exhibit the phenomenon of feature distribution shift when directly applying existing PEFTs to graph-based tasks. Further, our study empirically shows that the graph structure and Bregman proximal point optimization could alleviate this concern well.
    \item We propose a structure-aware parameter-efficient method (G-Adapter) for adapting pre-trained GTNs to various graph-based downstream tasks, in which G-Adapter introduces graph structure as an inductive bias to guide the updating process.
    \item Extensive experiments demonstrate that G-Adapter outperforms the counterparts by a significant margin. Furthermore, compared to full fine-tuning, our method yields tremendous memory footprint benefits almost without sacrificing the efficiency of training and inference.
\end{itemize}

\section{Related Work}
\paragraph{Graph Transformer Networks.}
Transformer \cite{vaswani2017attention}, as one of the most popular network architectures so far, has demonstrated remarkable success in NLP \cite{brown2020language, devlin2019bert, lewis2020bart, liu2019roberta, raffel2020exploring, touvron2023llama}, which spurs extensive research on transferring Transformer to graph representation learning \cite{min2022transformer, xia2022survey}. Considering the intrinsic discrepancy between graph and text modalities, current efforts mainly focus on two aspects: \textit{the design of pre-training tasks} and \textit{the encoding of nodes and edges}. For the first aspect, there are generally three folds:
(\romannumeral 1) Supervised learning: the preset supervised task is constructed by measuring the labels of graph data using professional tools \cite{hu*2020strategies, rong2020selfsupervised, zhang2021motifbased}.
(\romannumeral 2) Graph autoregressive modeling: similar to the GPT-style pre-training tasks \cite{brown2020language, radford2018improving, radford2019language} in NLP, some nodes and edges in the graph are randomly masked first, and then the masked elements are recovered in a step-by-step manner \cite{hu2020gptgnn, zhang2021motifbased}.
(\romannumeral 3) Masked components modeling: this approach is analogous to the MLM task in BERT \cite{devlin2019bert}, where all masked elements in the graph are predicted simultaneously \cite{hu*2020strategies, rong2020selfsupervised}.
For the second aspect, each node (e.g., an atom in the molecular graph) is regarded as a ``token'' in text sequence, and then the hidden representation of the node is learned similar to Transformers in NLP \cite{chen2023nagphormer}. Compared to the simple sequential relationship between tokens in text sequence, the relationship between edges in the graph could be more complex and essential \cite{min2022transformer, xia2022survey}. Therefore, substantial works focus on modeling graph structures \cite{bo2023specformer, chen2022structureaware, diao2023relational, jin2023edgeformers, kim2022pure, kreuzer2022rethinking, mialon2021graphit, ying2021transformers, yuan2023monocular, zhao2021gophormer}.
For example, Graphormer \cite{ying2021transformers} leverages the centrality and spatial encoding as the graph structural signal, and MAT \cite{maziarka2020molecule} augments the attention mechanism in Transformer using inter-atomic distances and the molecular graph structure. \citet{kreuzer2022rethinking} propose the learnable structural encoding via Laplacian spectrum, which can learn the position of each node in the graph. Moreover, \citet{zhao2021gophormer} proposes a proximity-enhanced multi-head attention to capture the multi-hop graph structure, and \citet{khoo2020interpretable} design a structure-aware self-attention for modeling the tree-structured graphs. Additionally, \citet{min2022transformer} systematically investigate the effectiveness and application of Transformers in the graph domain.

\paragraph{Parameter-Efficient Transfer Learning.}
Parameter-Efficient Fine-Tuning (PEFT) is receiving considerably growing attention in diverse domains \cite{ding2022delta, he2022unified, pfeiffer2020adapterhub, yu2022unified}. Adapter \cite{houlsby2019parameterefficient}, as the representative work of PEFT, is proposed to tackle natural language understanding tasks by inserting the compact blocks into Multi-Head Attention (MHA) and Feed-Forward Networks (FFN) in Transformer. Following this work, a series of subsequent efforts are proposed to improve the performance of Adapter. For instance, AdapterDrop \cite{ruckle2021adapterdrop} removes Adapter blocks from the lower layers, and Compacter/Compacter++ \cite{mahabadi2021compacter} introduce Kronecker product and weights sharing tricks to further reduce the proportion of trainable parameters. More similar works are included \cite{fu2022adapterbias, karimimahabadi2021parameterefficient, pfeiffer2021adapterfusion, stickland2019bert, wang2021kadapter, wang2022adamix}. Based on the hypothesis of \textit{low intrinsic rank} \cite{aghajanyan2021intrinsic}, LoRA \cite{hu2022lora} tunes two low-rank learnable matrices to approximate the updating of query and value weights in MHA. Moreover, \citet{zhang2023adaptive} enhance LoRA by adaptively allocating the trainable parameters budget at each layer, and FacT \cite{jie2022fact} extends LoRA by introducing a new tensorization-decomposition framework. Instead of introducing extra parameters, BitFit \cite{ben-zaken2022bitfit}, a simple heuristic strategy, only fine-tunes the bias terms of the model. In addition, prompt-based tuning \cite{lester2021power, li2021prefixtuning, liu2022ptuning, vu2022spot, wang2023multitaska} is also an interesting direction, but we do not involve these methods here given that \textit{the training curse of prompt-based methods} \cite{ding2022delta}. In addition, substantial works attempt to combine different PEFTs together through tailored mechanisms \cite{chen2023parameterefficient, hu2022sparse, jiang2023rethinking, mao2022unipelt, zhou2023autopeft}. Finally, \citet{he2022unified} provide a unified view of existing PEFTs, and more detailed descriptions of PEFTs are discussed in the survey literature \cite{ding2022delta}.

\section{Methodology}
\subsection{Pilot Experiments}
\label{subsec:pilot_exp}
To answer the first question: Can PEFTs from the language domain be transferred directly to graph-based tasks? We evaluate the performance of three mainstream PEFTs (Adapter \cite{houlsby2019parameterefficient}, LoRA \cite{hu2022lora}, and BitFit \cite{ben-zaken2022bitfit}) on large- and small-scale graph-based downstream tasks, respectively. To be specific, on the large-scale datasets, i.e., MolHIV ($41$K) and MolPCBA ($437$K) \cite{hu2021open}, we first average the results of each PEFT based on the pre-trained Graphormer \cite{ying2021transformers}, and then subtract the average result of full fine-tuning. Here, we refer to the final result as \textit{Performance Gap}, as shown in Fig.~\ref{fig:intro_compare}. Similar operations are also conducted on seven small-scale datasets ($0.6 \sim 2.4$K), i.e., FreeSolv, ESOL, BBBP, Estrogen-$\alpha$, Estrogen-$\beta$, MetStab$_\mathrm{low}$, and MetStab$_\mathrm{high}$ \cite{gaulton2012chembl, podlewska2018metstabon, wu2018moleculenet}, based on another pre-trained model MAT \cite{maziarka2020molecule}. From the comparison in Fig.~\ref{fig:intro_compare}, we can observe that the performance of traditional PEFTs is far from full fine-tuning on graph-based tasks, especially on large-scale datasets, across varying degrees of the ratio of trainable parameters.

To shed light on why there is such a significant gap between traditional PEFTs and full fine-tuning, we investigate the feature distribution of different methods inspired by \citet{lian2022scaling}. Specifically, based on BBBP and pre-trained MAT, we first take the hidden representation of a virtual node (similar to the \texttt{[CLS]} token in NLP \cite{devlin2019bert}) in the last layer as the entire graph representation. Then, for the identical input, the graph feature representations from diverse methods are visualized in Fig.~\ref{fig:feat_dist_shift}. More results are shown in Appendix~\ref{app:exp}. Given that full fine-tuning updates all parameters of the model, its performance can be seen as an ``upper bound'' for PEFT\footnote{Note that this claim is not rigorous, since PEFTs might outperform full fine-tuning on small-scale datasets.}. Therefore, a good PEFT is believed that it should have similar behavior with full fine-tuning, such as the encoding of features. However, from the comparison in Fig.~\ref{fig:feat_dist_shift}, we can observe that the feature distributions encoded by traditional PEFTs are shifted compared to full fine-tuning, which here is called \textit{feature distribution shift}.

To understand the reason underlying this phenomenon, we revisit the relationship between GTNs and vanilla Transformers. For the encoding of node/token, they have highly similar operations, e.g., encoding the representation of node/token through an embedding layer. However, there are significant discrepancies in terms of the encoding of position (or edge in the graph). Specifically, only the position embedding layer is utilized within vanilla Transformers in NLP, while most existing GTNs extract diverse graph structure information as the inductive bias and then inject them into the model \cite{maziarka2020molecule, rong2020selfsupervised, ying2021transformers, zhang2020graphbert, zhao2021gophormer}, since the graph structure contains rich edge semantic information. In addition, recent researches also demonstrate the significant effectiveness of graph structure in learning graph representation \cite{chen2022structureaware, diao2023relational, kreuzer2022rethinking}. Motivated by these observations, in this paper, we attempt to introduce graph structure as the inductive bias to alleviate the feature distribution shift issue.

\subsection{Structure-Aware Parameter-Efficient Fine-Tuning}
For the parameter-efficient module, we believe that the following principles should be taken into consideration:
(\romannumeral 1) it can explicitly encode graph structure during the fine-tuning process;
(\romannumeral 2) it should satisfy the main property of PEFT --- lightweight \cite{houlsby2019parameterefficient, hu2022lora, mahabadi2021compacter};
(\romannumeral 3) it should be easy to implement and can be integrated into diverse GTNs.

\paragraph{The design of parameter-efficient module.}
Inspired by the design of Graph Convolutional Networks (GCN) \cite{kipf2017semisupervised, lin2021mesh}, which can model both graph structure and node representation simultaneously, we leverage this operation to explicitly introduce graph structure into the model. Here, we give the following definition:
\begin{equation}
    X' = \mathrm{GraphConv}(S, X; W) = \sigma(S X W)
\end{equation}
where $X, X' \in \mathbb{R}^{n \times d}$ ($n$: the sequence length, $d$: the hidden representation dimension) refer to the input, output of the module, respectively. $S \in \mathbb{R}^{n \times n}$ indicates the introduced graph structure information (e.g., the adjacency matrix of graph), $W \in \mathbb{R}^{d \times d}$ is the learnable weight, and $\sigma(\cdot)$ indicates the nonlinear activation function. Further, following the lightweight principle, we decompose $W$ into two low-rank matrices to reduce the number of learnable parameters, i.e., $W = W_{down} W_{up}$, where $W_{down} \in \mathbb{R}^{d \times r}, W_{up} \in \mathbb{R}^{r \times d}$ and $r$ is called the bottleneck size. Moreover, to stabilize the training process of PEFT, we insert two LayerNorm layers \cite{ba2016layer} before and after $\mathrm{GraphConv}(\cdot)$, respectively, as depicted in Fig.~\ref{fig:overview}.

Overall, the pipeline of our PEFT module is as follows: firstly, the input ($X$) goes through the first LayerNorm layer, then passes $\mathrm{GraphConv}(\cdot)$ by absorbing the graph structure information ($S$). Next, we construct a skip connection between the output of $\mathrm{GraphConv}(\cdot)$ and the normalized input ($X'$). Lastly, the final output ($X''$) is obtained through the second LayerNorm layer, i.e.:
\begin{equation}
    X' = \mathrm{LN}(X), X'' = \mathrm{LN} \big( X' + \sigma ( S X' W_{down} W_{up} ) \big)
\end{equation}
where $\mathrm{LN}(\cdot)$ represents the LayerNorm layer. In Fig.~\ref{fig:overview}, we describe in detail the architecture of our proposed PEFT (G-Adapter) and compare it with traditional PEFTs. In addition, thanks to the modular and lightweight properties, G-Adapter can be seamlessly integrated into diverse GTNs. We also provide some general pseudo-code to execute our approach in Appendix~\ref{app:config}.

\paragraph{The selection of graph structure information.}
To start with, we consider the adjacency matrix (with self-connections) of the graph: $S_{1} = A + I_{n}$, where $A, I_{n} \in \mathbb{R}^{n \times n}$ refer to the adjacency matrix and identity matrix, respectively. Then, following \citet{kipf2017semisupervised}, we introduce the degree matrix of nodes to normalize the adjacency matrix: $S_{2} = \tilde{D}^{-\frac{1}{2}} \tilde{A} \tilde{D}^{-\frac{1}{2}}$, where $\tilde{A} = S_{1}$, $\tilde{D}$ is the diagonal matrix and $\tilde{D}_{ii} = \sum_{j} \tilde{A}_{ij}$. In addition, we propose a distance-based graph structure information: $S_{3} = [\mathrm{dis}(v_{i}, v_{j})]_{n \times n}$, where $\mathrm{dis}(v_{i}, v_{j})$ refers to the distance of the shortest path (or the inter-atomic distance in the molecular graph) between two nodes $v_{i}$ and $v_{j}$. Last, we combine the adjacency and distance to construct a hybrid structure information: $S_{4} = \alpha \cdot \tilde{D}^{-\frac{1}{2}} \tilde{A} \tilde{D}^{-\frac{1}{2}} + \beta \cdot [\mathrm{dis}(v_{i}, v_{j})]_{n\times n}$, where $\alpha, \beta$ are scalar hyper-parameters to balance the impacts of the adjacency and distance terms. We evaluate the proposed graph structure information ($S_{1}, S_{2}, S_{3}, S_{4}$) on a range of graph-based tasks, and the detailed comparisons are presented in Sec.~\ref{subsec:main_res}.

\begin{wrapfigure}{r}{0.4\textwidth}
    \vspace{-1.5em}
    \centering
    \includegraphics[width=0.4\textwidth]{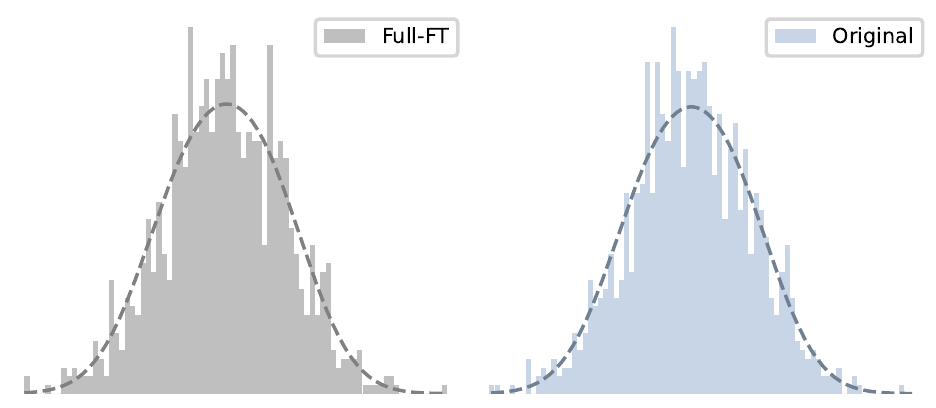}
    \caption{Comparison of feature distribution between full fine-tuning and the original model parameters, where Jensen-Shannon divergence is $0.27\%$.}
    \label{fig:feat_compare}
    \vspace{-2.0em}
\end{wrapfigure}

\subsection{Bregman Proximal Point Optimization}
It is expected that the feature distribution encoded by PEFT should be aligned with full fine-tuning as much as possible, as discussed in Sec.~\ref{subsec:pilot_exp}. However, the feature distribution of full fine-tuning is unavailable during the training process of PEFT. Interestingly, we observe that the feature distribution encoded by the original model parameters has a high similarity with full fine-tuning, as shown in Fig.~\ref{fig:feat_compare}. It indicates that full fine-tuning only slightly modulates the value of model parameters but does not change the ability of the model to encode features.

Therefore, to maintain consistency with the feature distribution of the original parameters, we propose Bregman proximal point optimization strategy \cite{jiang2020smart} to prevent the model from aggressive update. Specifically, for the pre-trained model $f(\cdot; \theta)$ with trainable parameters $\theta$, at the $(t + 1)$-th iteration, we have
\begin{equation}
    \theta_{t + 1} = \arg\min_{\theta} ~ (1 - \mu) \cdot \mathcal{L}_{\mathrm{vanilla}}(\theta) + \mu \cdot \mathcal{L}_{\mathrm{bregman}}(\theta, \theta_{t})
\end{equation}
where $\mu > 0$ is a hyper-parameter, $\mathcal{L}_{\mathrm{vanilla}}$ is a common classification or regression loss function, and $\mathcal{L}_{\mathrm{bregman}}$ is the Bregman divergence defined as:
\begin{equation}
    \mathcal{L}_{\mathrm{bregman}}(\theta, \theta_{t}) = \mathbb{E}_{x \sim \mathcal{D}} \Big[\ell \big(f(x; \theta), f(x; \theta_{t})\big)\Big]
\end{equation}
where the input $x$ is derived from the training set $\mathcal{D}$, and here we leverage the symmetric KL-divergence, i.e., $\ell(p, q) = \mathrm{KL}(p||q) + \mathrm{KL}(q||p)$. Intuitively, $\mathcal{L}_{\mathrm{bregman}}$ serves as a regularizer and prevents $\theta_{t + 1}$ from deviating too much from the previous iteration $\theta_{t}$, therefore can effectively retain the capacity of encoding feature in the pre-trained model $f(\cdot; \theta)$.

\section{Experiments}
\subsection{Setup}
\label{subsec:setup}

\paragraph{Datasets \& Evaluation Protocols.}
We conduct our experiments on nine benchmark datasets: MolHIV, MolPCBA, FreeSolv, ESOL, BBBP, Estrogen-$\alpha$, Estrogen-$\beta$, MetStab$_\mathrm{low}$ and MetStab$_\mathrm{high}$ \cite{gaulton2012chembl, hu2021open, podlewska2018metstabon, wu2018moleculenet}, where MolHIV ($41$K) and MolPCBA ($437$K) are two large-scale molecular property prediction datasets and the others are small-scale molecular datasets ($0.6 \sim 2.4$K). We provide more descriptions of datasets in Appendix~\ref{app:datasets}. Following the previous settings \cite{maziarka2020molecule, ying2021transformers}, we employ the scaffold split on MolHIV, MolPCBA, BBBP, and Estrogen-$\alpha/\beta$, and then the random split on the rest of datasets. For the evaluation protocols, MolPCBA is evaluated by Accuracy Precision (AP), FreeSolv and ESOL are evaluated by RMSE, and the others are evaluated by AUC.

\begin{table}[t]
    \centering
    \caption{Comparison of PEFTs and full fine-tuning on small-scale datasets. The results are averaged from six seeds, and the subscript is the standard deviation, where \textbf{bold} indicates the best results in PEFTs. $\ast$ represents the mean ratio of trainable parameters over seven datasets.}
    \scalebox{0.72}{
        \begin{tabular}{lcccccccc}
            \toprule
                \multirow{2}{*}{Method} & \multirow{2}{*}{Ratio$^{\ast}$} & \multicolumn{2}{c}{RMSE ($\downarrow$)} & \multicolumn{5}{c}{AUC ($\uparrow$)} \\
            \cmidrule(l){3-4} \cmidrule(l){5-9}
                 & ($\gamma$) & FreeSolv & ESOL & BBBP & Estrogen-$\alpha$ & Estrogen-$\beta$ & MetStab$_\mathrm{low}$ & MetStab$_\mathrm{high}$ \\
            \midrule
             Full Finetunig & $100\%$  & $0.286_{\pm 0.035}$ & $0.270_{\pm 0.037}$ & $0.764_{\pm 0.008}$ & $0.979_{\pm 0.002}$ & $0.778_{\pm 0.005}$ & $0.863_{\pm 0.025}$ & $0.878_{\pm 0.032}$ \\
            \midrule
                Adapter     & $2.52\%$ & $0.327_{\pm 0.011}$ & $0.320_{\pm 0.072}$ & $0.724_{\pm 0.009}$ & $0.978_{\pm 0.024}$ & $0.768_{\pm 0.021}$ & $0.846_{\pm 0.034}$ & $0.859_{\pm 0.028}$ \\
                
                Hyperformer & $2.43\%$ & $0.310_{\pm 0.020}$ & $0.321_{\pm 0.045}$ & $0.727_{\pm 0.012}$ & $0.977_{\pm 0.027}$ & $0.770_{\pm 0.013}$ & $0.842_{\pm 0.022}$ & $0.853_{\pm 0.023}$ \\
                Compacter   & $1.56\%$ & $0.314_{\pm 0.028}$ & $0.316_{\pm 0.038}$ & $0.730_{\pm 0.022}$ & $0.971_{\pm 0.034}$ & $0.764_{\pm 0.027}$ & $0.832_{\pm 0.019}$ & $0.860_{\pm 0.046}$ \\
                MAM         & $1.28\%$ & $0.302_{\pm 0.019}$ & $0.292_{\pm 0.022}$ & $0.743_{\pm 0.014}$ & $0.980_{\pm 0.011}$ & $0.776_{\pm 0.022}$ & $0.851_{\pm 0.023}$ & $0.872_{\pm 0.054}$ \\
                
                LoRA        & $1.01\%$ & $0.309_{\pm 0.032}$ & $0.284_{\pm 0.054}$ & $0.726_{\pm 0.012}$ & $0.979_{\pm 0.007}$ & $0.781_{\pm 0.039}$ & $0.839_{\pm 0.022}$ & $0.878_{\pm 0.027}$ \\
                BitFit      & $0.10\%$ & $0.321_{\pm 0.048}$ & $0.314_{\pm 0.031}$ & $0.739_{\pm 0.005}$ & $0.977_{\pm 0.019}$ & $0.770_{\pm 0.035}$ & $0.848_{\pm 0.031}$ & $0.805_{\pm 0.045}$ \\
            \midrule
                G-Adapter ($S_{1}$) & $0.71\%$ & $\textbf{0.280}_{\pm 0.012}$ & $\textbf{0.279}_{\pm 0.018}$ & $0.750_{\pm 0.012}$ & $0.976_{\pm 0.033}$ & $\textbf{0.791}_{\pm 0.022}$ & $0.865_{\pm 0.036}$ & $\textbf{0.881}_{\pm 0.023}$ \\
                G-Adapter ($S_{2}$) & $0.71\%$ & $0.282_{\pm 0.014}$ & $0.286_{\pm 0.022}$ & $\textbf{0.751}_{\pm 0.009}$ & $\textbf{0.981}_{\pm 0.017}$ & $0.788_{\pm 0.031}$ & $\textbf{0.870}_{\pm 0.013}$ & $0.874_{\pm 0.025}$ \\
                G-Adapter ($S_{3}$) & $0.71\%$ & $0.291_{\pm 0.008}$ & $0.289_{\pm 0.017}$ & $0.744_{\pm 0.011}$ & $0.973_{\pm 0.015}$ & $0.786_{\pm 0.034}$ & $0.860_{\pm 0.031}$ & $0.861_{\pm 0.018}$ \\
                G-Adapter ($S_{4}$) & $0.71\%$ & $0.298_{\pm 0.011}$ & $0.282_{\pm 0.019}$ & $0.747_{\pm 0.006}$ & $0.975_{\pm 0.011}$ & $0.775_{\pm 0.024}$ & $0.858_{\pm 0.025}$ & $0.869_{\pm 0.037}$ \\
            \bottomrule
        \end{tabular}
    }
    \label{tab:small_scale}
    \vspace{-1.0em}
\end{table}

\paragraph{Pre-trained Models \& Baselines.}
Two widely used pre-trained GTNs are leveraged as our backbones: Graphormer \cite{ying2021transformers} and MAT \cite{maziarka2020molecule}. In our experiments, we employ the base version of Graphormer, which has $12$ layers Transformer encoders and is pre-trained on large-scale molecular dataset PCQM4M-LSC \cite{hu2021ogblsc}. MAT is built on $8$ encoders of Transformer, where the dimension of hidden representation is set to $1024$. And, the node-level self-supervised learning serves as a pre-training task for MAT on ZINC15 \cite{sterling2015zinc}. For the baselines, we include full fine-tuning as a strong counterpart and six popular traditional PEFTs: Adapter \cite{houlsby2019parameterefficient}, LoRA \cite{hu2022lora}, BitFit \cite{ben-zaken2022bitfit}, Hyperformer \cite{karimimahabadi2021parameterefficient}, Compacter \cite{mahabadi2021compacter}, and MAM \cite{he2022unified}. More descriptions per baseline are presented in Appendix~\ref{app:pre}.

\paragraph{Implementation.}
Before fine-tuning, we begin by reusing the official released pre-trained checkpoints\footnote{\url{https://github.com/microsoft/Graphormer}; \url{https://github.com/ardigen/MAT}} to initialize our backbones, while the introduced modules are randomly initialized, and then use AdamW optimizer to fine-tune the models. We set fair hyperparametric search budgets for various PEFTs, and the detailed configurations per method on diverse datasets are shown in Appendix~\ref{app:config}.

\begin{wraptable}{r}{0.5\textwidth}
    \vspace{-2.0em}
    \centering
    \caption{The comparison on two large-scale datasets MolHIV and MolPCBA.}
    \scalebox{0.66}{
        \begin{tabular}{lcccc}
            \toprule
                \multirow{2}{*}{Method} & \multicolumn{2}{c}{MolHIV} & \multicolumn{2}{c}{MolPCBA} \\
            \cmidrule(l){2-3} \cmidrule(l){4-5}
                & Ratio ($\gamma$) & AUC ($\uparrow$) & Ratio ($\gamma$) & AP ($\uparrow$) \\
            \midrule
                Full Finetunig & $100\%$ & $0.804_{\pm 0.006}$ & $100\%$ & $0.272_{\pm 0.013}$ \\
            \midrule
                Adapter     & $1.24\%$ & $0.743_{\pm 0.010}$ & $4.69\%$ & $0.235_{\pm 0.009}$ \\
                
                Hyperformer & $1.13\%$ & $0.740_{\pm 0.012}$ & $4.37\%$ & $0.246_{\pm 0.012}$ \\
                Compacter   & $0.64\%$ & $0.752_{\pm 0.023}$ & $3.42\%$ & $0.230_{\pm 0.023}$ \\
                MAM         & $0.57\%$ & $0.758_{\pm 0.017}$ & $2.66\%$ & $0.251_{\pm 0.016}$ \\
                
                LoRA        & $0.34\%$ & $0.763_{\pm 0.014}$ & $2.42\%$ & $0.246_{\pm 0.012}$ \\
                BitFit      & $0.16\%$ & $0.709_{\pm 0.008}$ & $0.16\%$ & $0.184_{\pm 0.011}$ \\
            \midrule
                G-Adapter ($S_{1}$) & $0.24\%$ & $\textbf{0.790}_{\pm 0.011}$ & $1.89\%$ & $\textbf{0.269}_{\pm 0.008}$ \\
                G-Adapter ($S_{2}$) & $0.24\%$ & $0.788_{\pm 0.006}$ & $1.89\%$ & $0.264_{\pm 0.012}$ \\
                G-Adapter ($S_{3}$) & $0.24\%$ & $0.772_{\pm 0.008}$ & $1.89\%$ & $0.250_{\pm 0.011}$ \\
                G-Adapter ($S_{4}$) & $0.24\%$ & $0.781_{\pm 0.012}$ & $1.89\%$ & $0.262_{\pm 0.011}$ \\
            \bottomrule
        \end{tabular}
    }
    \label{tab:large_scale}
    \vspace{-1.5em}
\end{wraptable}

\subsection{Main Results}
\label{subsec:main_res}
We report the comparison results on MolHIV and MolPCBA based on the pre-trained Graphormer in Tab.~\ref{tab:large_scale}, and more results are shown in Tab.~\ref{tab:small_scale} on small-scale datasets based on the pre-trained MAT. From these comparisons, we can draw the following observations:

\textbf{Observation \uppercase\expandafter{\romannumeral 1}}:
Simple graph structure could deliver significant performance benefits. For instance, based on graph adjacent information, G-Adapters ($S_{1}, S_{2}$) obtain better results compared to G-Adapters ($S_{3}, S_{4}$) with distance-based structure information in Tab.~\ref{tab:small_scale}. Moreover, in Tab.~\ref{tab:large_scale}, G-Adapter ($S_{1}$) achieves the optimal performance among all PEFTs. We speculate that this may be because the graph adjacency matrix ($S_{1}$) is the complete information of graph structures, that is, the distance-based structure ($S_{3}$) can be derived from $S_{1}$. Therefore, this suggests that our model is not only remarkably expressive but also insensitive to the graph structure, which means that we do not have to design tailored graph structures except for graph adjacency information. In the following statement, we take G-Adapter ($S_{1}$) as our baseline unless otherwise specified.

\textbf{Observation \uppercase\expandafter{\romannumeral 2}}:
Our proposed G-Adapter consistently outperforms traditional PEFTs and offers a better trade-off between the ratio of trainable parameters ($\gamma$) and performance. For instance, Adapter, LoRA, and BitFit lag far behind G-Adapter on MolHIV and MolPCBA in Tab.~\ref{tab:large_scale}. Although BitFit updates the fewest number of parameters ($\gamma = 0.16\%$), it also yields the worst performances ($0.709$ AUC, $0.184$ AP). In comparison, our proposed G-Adapter achieves $79.0$ AUC, $0.269$ AP with $\gamma = 0.24\%, \gamma = 1.89\%$, respectively, which is also the optimal solution compared to other PEFTs.

\textbf{Observation \uppercase\expandafter{\romannumeral 3}}:
Compared to full fine-tuning, G-Adapter could achieve competitive and (most) superior performances on large- and small-scale datasets, respectively. For example, G-Adapter outperforms full fine-tuning by a significant margin ($1.3\%$ AUC) on Estrogen-$\beta$. We believe that the improvement on small-scale datasets is understandable, there are several reasons:
(\romannumeral 1) With decreasing the scale of the training set, it might meet serious over-fitting and catastrophic forgetting issues if the entire parameters are updated in full fine-tuning, whereas our method eases these concerns by only tuning G-Adapter blocks while freezing the original parameters.
(\romannumeral 2) G-Adapter restricts the drastic updating of parameters via Bregman proximal point optimization strategy, which acts as a regularizer during the training process and therefore boosts the generalization capacity.

\begin{figure}[t]
    \centering
    \subfigure[MolPCBA ($437$K)]{
        \includegraphics[width=0.3\textwidth]{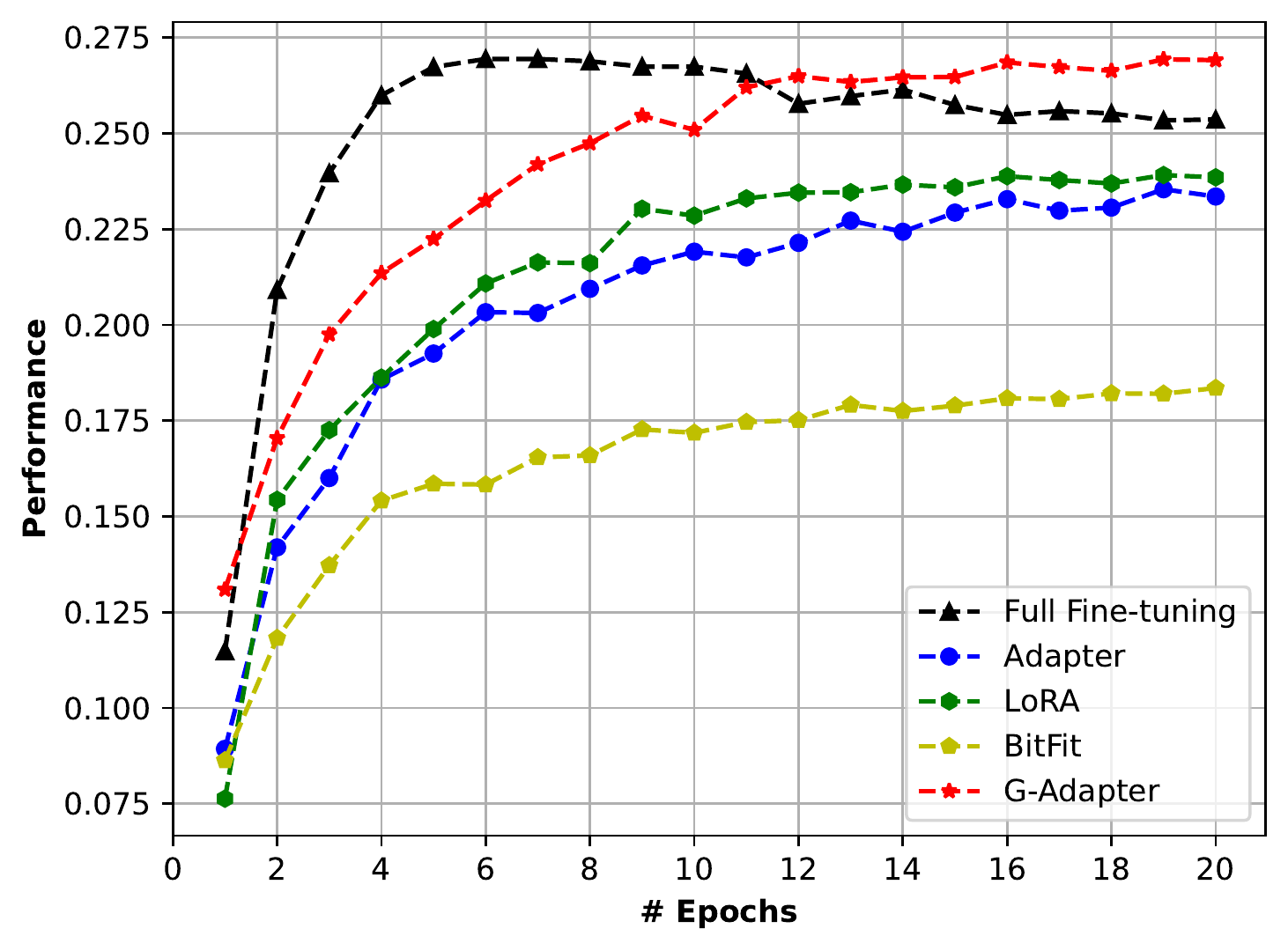}
    }
    \subfigure[MolHIV ($41$K)]{
        \includegraphics[width=0.3\textwidth]{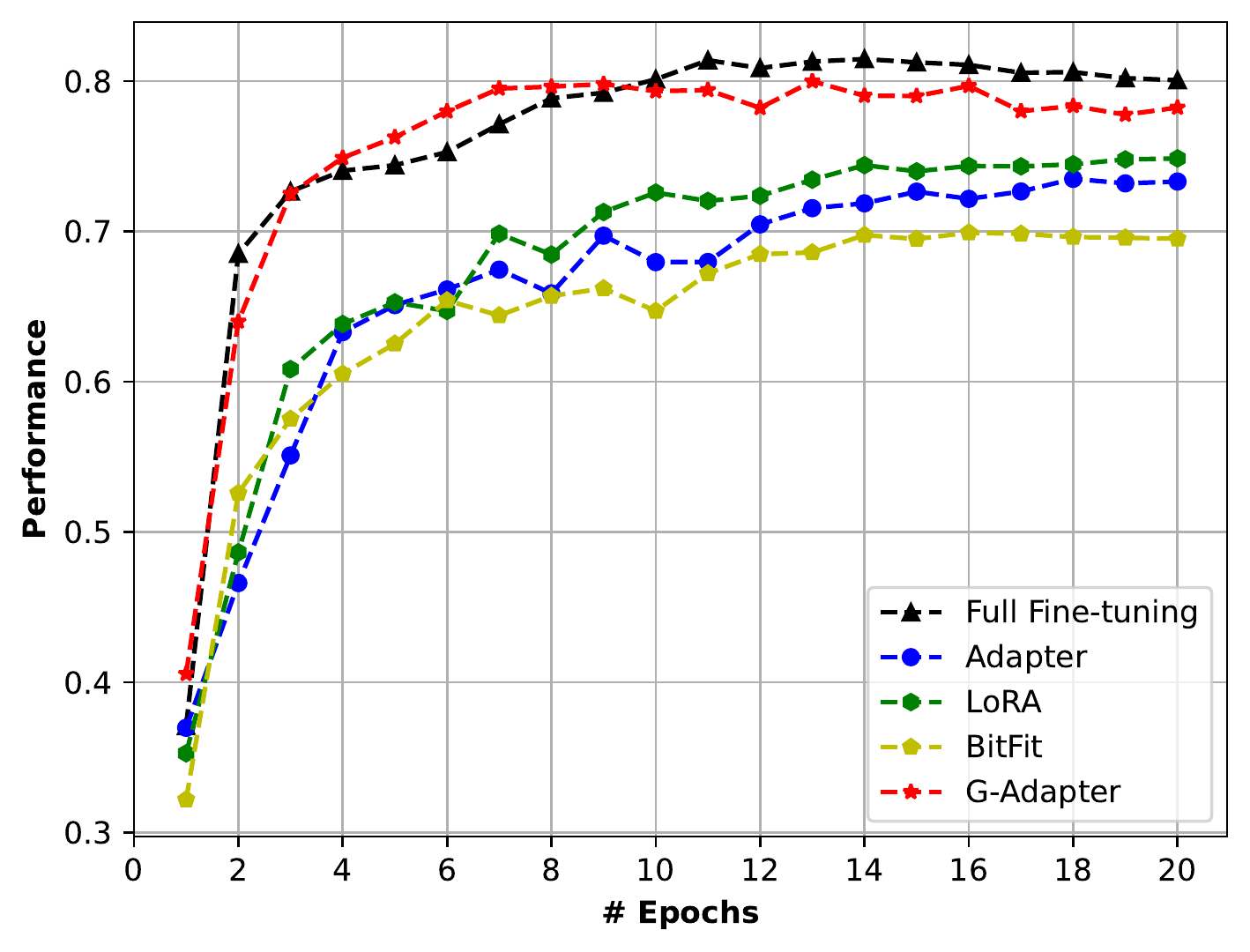}
    }
    \subfigure[Estrogen-$\alpha$ ($2$K)]{
        \includegraphics[width=0.3\textwidth]{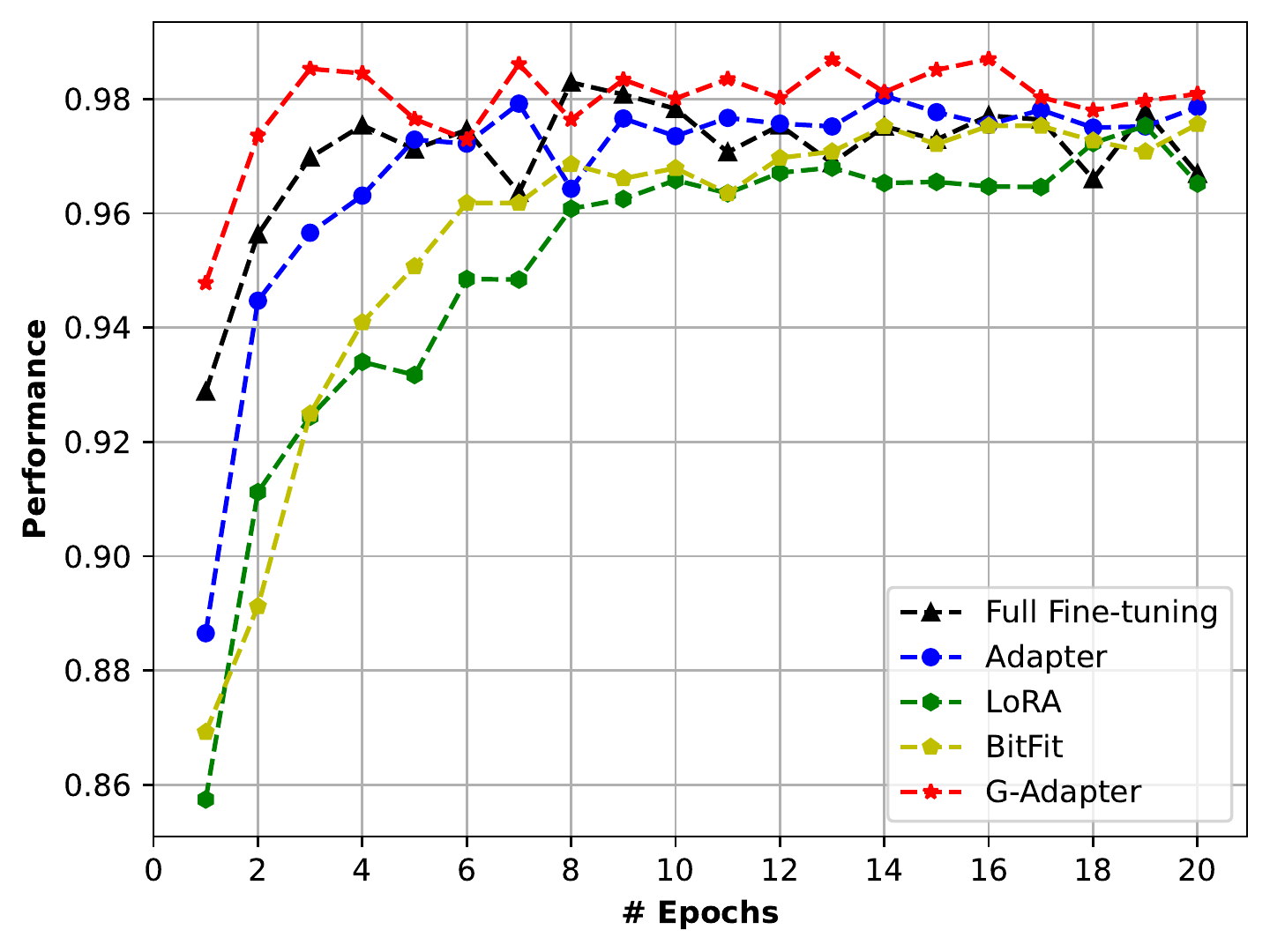}
    }
    \caption{The comparison of training efficiency between PEFTs (Adapter, LoRA, BitFit, and our proposed G-Adapter) and full fine-tuning on diverse scale datasets.}
    \label{fig:compare_train_eff}
\end{figure}

\section{Analysis}

\begin{table}[t]
    \centering
    \caption{The comparison of PEFTs and full fine-tuning in terms of inference speed (the millisecond per sample) and memory footprint across different bottleneck sizes.}
    \scalebox{0.76}{
        \begin{tabular}{lcccccccc}
            \toprule
                \multirow{2}{*}{Method} & \multicolumn{2}{c}{MolPCBA$_{(r=32)}$} & \multicolumn{2}{c}{MolHIV$_{(r=16)}$} & \multicolumn{2}{c}{FreeSolv$_{(r=8)}$} & \multicolumn{2}{c}{Estrogen-$\alpha$$_{(r=4)}$}  \\
            \cmidrule(l){2-3} \cmidrule(l){4-5} \cmidrule(l){6-7} \cmidrule(l){8-9}
                & Mem. (MB) & Infer (ms) & Mem. (MB) & Infer (ms) & Mem. (MB) & Infer (ms) & Mem. (MB) & Infer (ms) \\
            \midrule
                Full Finetunig & $185$ & $0.81$ & $185$ & $1.39$ & $161$ & $0.42$ & $161$ & $1.11$ \\
            \midrule
                Adapter & $5.0$ & $0.99$ & $2.4$ & $1.46$ & $1.1$ & $0.46$ & $0.7$ & $1.15$ \\
                LoRA    & $5.0$ & $0.97$ & $2.4$ & $1.43$ & $1.1$ & $0.47$ & $0.7$ & $1.13$ \\
                BitFit  & $0.3$ & $0.81$ & $0.3$ & $1.39$ & $0.2$ & $0.42$ & $0.2$ & $1.11$ \\
            \midrule
              G-Adapter & $2.9$ & $0.93$ & $1.4$ & $1.41$ & $0.7$ & $0.44$ & $0.4$ & $1.12$ \\
            \bottomrule
        \end{tabular}
    }
    \label{tab:eff_inf_mem}
    \vspace{-1.5em}
\end{table}

\subsection{Efficiency of Training, Inference and Memory Footprint}
\label{subsec:eff_tra_inf_mem}
In this subsection, we mainly investigate the following questions: Does G-Adapter seriously affect the training (or convergence) efficiency and inference speed compared to full fine-tuning? And, can G-Adapter bring significant benefits to the storage of model weights, as we claimed before? Specifically, for the training efficiency, we evaluate PEFTs and full fine-tuning on three datasets with different scales\footnote{The experiments are conducted on single NVIDIA GeForce RTX 3090 GPU (24G), where CPU is AMD EPYC 7763 64-Core Processor.}: MolPCBA ($437$K), MolHIV ($41$K), and Estrogen-$\alpha$ ($2$K). The experimental results are shown in Fig.~\ref{fig:compare_train_eff}, in which we observe that:
(\romannumeral 1) For the large-scale dataset, the convergence of PEFTs lags behind full fine-tuning by about $4 \sim 5$ epochs. However, this gap is significantly narrowed as the amount of training data decreases.
(\romannumeral 2) Compared to traditional PEFTs, G-Adapter not only achieves faster convergence but also higher performance over datasets of varying scales.

For the inference efficiency and memory footprint, we adopt different bottleneck sizes ($r = 32, 16, 8, 4$) on MolPCBA, MolHIV, FreeSolv, and Estrogen-$\alpha$, respectively. The experimental results are shown in Tab.~\ref{tab:eff_inf_mem}, where we observe that:
(\romannumeral 1) Compared to full fine-tuning, the extra introduced modules result in a trivial inference delay, which is almost negligible with the bottleneck size decreasing. Note that BitFit does not introduce additional modules but merely tunes the bias terms, therefore it theoretically has the same inference efficiency as full fine-tuning.
(\romannumeral 2) For the storage requirements, the distinction between PEFTs and full fine-tuning is remarkably significant. For an example on Estrogen-$\alpha$ ($r = 4$), full fine-tuning requires storing a complete checkpoint ($161$MB) for each downstream task, while Adapter (LoRA), BitFit and G-Adapter only need to store $0.7$MB, $0.2$MB and $0.4$MB checkpoint per task, respectively, which greatly reduces the storage requirements.

\subsection{The Impact of Insertion Position and Components}
\label{subsec:impact_design}
We discuss the impact of potential designs on performance from two perspectives:
(\romannumeral 1) The insertion position. First, we insert G-Adapter block into the front and back of MHA, denoted as \textbf{pre\_mha} and \textbf{post\_mha}, respectively. Similarly, G-Adapter block is plugged before and after FFN, denoted as \textbf{pre\_ffn} and \textbf{post\_ffn}, respectively. Note that our baseline can be regarded as inserting the G-Adapter block in the middle of FFN. Finally, like the insertion position of Adapter in Fig.~\ref{fig:overview}, we insert two G-Adapter blocks into MHA and FFN, denoted as \textbf{mha + ffn}.
(\romannumeral 2) Importance of each component. First, we remove the adjacency matrix (denoted as \textbf{w/o. S}) to explore the importance of graph structure. Then, we separately remove the first, second, and both LayerNorm layers to explore individual effects denoted as \textbf{w/o. pre\_ln}, \textbf{w/o. post\_ln} and \textbf{w/o. ln}. We also explore the role of nonlinear activation function by removing it, denoted as \textbf{w/o. act\_fn}. In addition, the effect of Bregman proximal point optimization is evaluated by only using the vanilla loss function, denoted as \textbf{w/o. breg}.

\begin{wraptable}{r}{0.4\textwidth}
    \vspace{-2.0em}
    \centering
    \caption{The impact of insertion position and components on performance.}
    \scalebox{0.7}{
        \begin{tabular}{lcc}
            \toprule
                Method & MolHIV & MolPCBA \\
            \midrule
                G-Adapter & $0.790$ & $0.269$ \\
            \midrule
                G-Adapter (pre\_mha) & $0.752$ & $0.246$ \\
                G-Adapter (post\_mha) & $0.747$ & $0.232$ \\
                G-Adapter (pre\_ffn) & $0.781$ & $0.258$ \\
                G-Adapter (post\_ffn) & $0.770$ & $0.260$ \\
                G-Adapter (mha + ffn) & $0.763$ & $0.245$ \\
            \midrule
                G-Adapter (w/o. $S$) & $0.728$ & $0.214$ \\
                G-Adapter (w/o. pre\_ln) & $0.762$ & $0.247$ \\
                G-Adapter (w/o. post\_ln) & $0.755$ & $0.250$ \\
                G-Adapter (w/o. ln) & $0.745$ & $0.237$ \\
                G-Adapter (w/o. act\_fn) & $0.766$ & $0.240$ \\
                G-Adapter (w/o. breg) & $0.754$ & $0.234$ \\
            \bottomrule
        \end{tabular}
    }
    \label{tab:impact_designs}
    \vspace{-0.5em}
\end{wraptable}

The experimental results are shown in Tab.~\ref{tab:impact_designs}, in which we can obtain that:
(\romannumeral 1) Plugging G-Adapter block into the front, middle, or back of FFN could yield better performance than MHA, and more blocks seem not to give better results, which is also consistent with the previous conclusion in NLP \cite{he2022unified}. An intuitive explanation is that, in each encoder of Transformer, FFN concentrates most of the parameters ($\sim 67\%$), while MHA only accounts for $\sim 33\%$. Therefore, tweaking the weights of FFN may be a more efficient way for fine-tuning.
(\romannumeral 2) Removing any of LayerNorm layers or the nonlinear activation function will hurt the performance. Moreover, removing the graph structure or Bregman proximal point optimization strategy would also significantly degrade the performance.

\begin{wraptable}{r}{0.4\textwidth}
    \vspace{-2.0em}
    \centering
    \caption{The impact of graph structure information on traditional PEFTs.}
    \scalebox{0.61}{
        \begin{tabular}{lcccc}
            \toprule
                Method & MolHIV & MolPCBA & BBBP & MetStab$_\mathrm{low}$ \\
            \midrule
                Adapter       & $0.743$ & $0.235$ & $0.724$ & $0.839$ \\
                Adapter + $S$ & $0.749$ & $0.242$ & $0.733$ & $0.844$ \\
            \midrule
                LoRA          & $0.763$ & $0.246$ & $0.739$ & $0.858$ \\
                LoRA + $S$    & $0.766$ & $0.252$ & $0.742$ & $0.861$ \\
            \bottomrule
        \end{tabular}
    }
    \label{tab:S_benefit}
    \vspace{-1.5em}
\end{wraptable}

\subsection{Can Graph Structure Information Benefit Traditional PEFTs?}
\label{subsec:can_graph}
One of the major contributions of G-Adapter is the introduction of graph structure, therefore a natural question is: can the graph structure enhance the traditional PEFTs as well? Given that the adjacency matrix ($S$) has performed well as graph structure information in previous experiments, we directly introduce $S$ into Adapter and LoRA. Their modified updating formulas are presented in Appendix~\ref{app:pre}. We conduct the experiments on four datasets: MolHIV, MolPCBA, BBBP, and MetStab$_\mathrm{low}$. The results are reported in Tab.~\ref{tab:S_benefit}, in which we could observe a slight improvement in the modified methods compared to the original Adapter and LoRA. However, there is still a significant gap with our proposed G-Adapter, which further justifies that the traditional PEFT architectures are not suitable for handling graph-based tasks.

\section{Conclusion \& Limitations}
In this paper, we propose a novel structure-aware PEFT method, G-Adapter, for graph-based tasks based on pre-trained GTNs. Unlike the traditional PEFTs, which lead to the issue of feature distribution shift, G-Adapter leverages the graph structure and Bregman proximal point optimization strategy to mitigate this concern. Extensive experiments on a variety of graph-based downstream tasks demonstrate the effectiveness of our proposed method. Although our approach demonstrates satisfactory performance, there are still some limitations:
(\romannumeral 1) Considering that the applicable scenarios of PEFT are large-scale models, our method is not tested on conventional graph network architectures (e.g., GCN \cite{kipf2017semisupervised}, GIN \cite{xu*2019how}). Because these models are already quite lightweight, resulting in the advantages of PEFT not being sufficiently exploited.
(\romannumeral 2) Limited by computational resources, we only evaluate two pre-trained GTNs (Graphormer and MAT). Nevertheless, thanks to the simplicity and generality of our proposed method, it can be applied to various graph Transformer-based models.

\clearpage
\bibliographystyle{ACM-Reference-Format}
\bibliography{reference}

\clearpage
\appendix
\section{Appendix}

\subsection{More Detailed Preliminaries}
\label{app:pre}
\paragraph{Transformer}
Transformer \cite{vaswani2017attention}, as one of the most popular network architectures so far, has been widely employed in diverse domains, such as NLP, computer vision and graph. In a standard encoder of Transformer, Multi-Head self-Attention (MHA) and Feed-Forward Networks (FFN) are two core components. Given the input $X \in \mathbb{R}^{n \times d}$, where $n$ is the length of input sequence and $d$ refers to the hidden size of representation, the query $Q$ and key-value pairs $K, V$ are first obtained by: $Q = XW_{q} + b_{q}, K = XW_{k} + b_{k}, V = XW_{v} + b_{v}$, where three projection matrices $W_{q/k/v} \in \mathbb{R}^{d \times d}$ and the bias $b_{q/k/v} \in \mathbb{R}^{d}$. Then, they are split into $N_{h}$ heads ($Q_{i}, K_{i}, V_{i} \in \mathbb{R}^{d \times d_{k}}, d_{k} = d / N_{h}$) to pass the self-attention operation:
\begin{equation}
    \mathrm{Attn}(Q_{i}, K_{i}, V_{i}) = \mathrm{softmax}\Big(\frac{Q_{i}K_{i}^{\top}}{\sqrt{d_{k}}}\Big)V_{i}
\end{equation}
After that, all head outputs are concatenated by a linear projection transformation ($W_{o} \in \mathbb{R}^{d \times d}, b_{o} \in \mathbb{R}^{d}$), then we can attain the final output of MHA:
\begin{equation}
    \mathrm{MAH}(X) = \mathrm{Concat}(\mathrm{head}_{1}, \mathrm{head}_{2}, \cdots, \mathrm{head}_{h})W_{o} + b_{o}
\end{equation}
where $\mathrm{head}_{i} = \mathrm{Attn}(Q_{i}, K_{i}, V_{i})$. Another important module is FFN, which consists of two linear layers with a ReLU nonlinear activation function (where we still take $X$ as the input for simplicity):
\begin{equation}
    \mathrm{FFN(X)} = \mathrm{ReLU}(XW_{f1} + b_{f1})W_{f2} + b_{f2}
\end{equation}
where $W_{f1} \in \mathbb{R}^{d \times d_{ff}}, W_{f2} \in \mathbb{R}^{d_{ff} \times d}, b_{f1} \in \mathbb{R}^{d_{ff}}, b_{f2} \in \mathbb{R}^{d}$. Note that $d_{ff} = d$ in some GTNs, such as Graphormer \cite{ying2021transformers} and MAT \cite{maziarka2020molecule}. In the following demonstration, for simplicity, we take $X, X' \in \mathbb{R}^{d \times d}$ as the input, output of a certain module, respectively.

\paragraph{Adapter} \citet{houlsby2019parameterefficient} insert two compact modules (i.e., Adapter blocks in Fig.~\ref{fig:overview}) into the encoders of Transformer. Specifically, an Adapter block is composed of the down-projection transformation $W_{down}$, the up-projection transformation $W_{up}$, the nonlinear activation function $\sigma(\cdot)$, and the skip connection:
\begin{equation}
    X' = X + \sigma(X W_{down}) W_{up}
    \label{eq:adapter}
\end{equation}
where $W_{down} \in \mathbb{R}^{d \times r}, W_{up} \in \mathbb{R}^{r \times d}$ and $r$ is the bottleneck size of Adapter, which satisfies the condition $r \ll d$ for reducing the number of learnable parameters. To introduce the graph structure information ($S \in \mathbb{R}^{n \times n}$) in Sec.~\ref{subsec:can_graph}, we modify Eq.~(\ref{eq:adapter}) as follows:
\begin{equation}
    X' = X + \sigma(S X W_{down}) W_{up}
    \label{eq:adapter_modify}
\end{equation}

\paragraph{LoRA}
Based on the \textit{low intrinsic rank} hypothesis \cite{aghajanyan2021intrinsic}, LoRA \cite{hu2022lora} reparametrizes the updating of pre-trained weight $W$ by the low-rank decomposition, i.e., $W + \Delta W = W + W_{down}W_{up}$. For the practice in Transformer, LoRA injects two low-rank modules into the query and value weights ($W_{q}, W_{v}$) in a parallel connection manner. For the weight $W_{*} \in \{W_{q}, W_{v}\}$, we can obtain:
\begin{equation}
    X' = X(W_{*} + s \cdot W_{down}W_{up})
    \label{eq:lora}
\end{equation}
where $s \ge 1$ is a scalar hyper-parameter, and $X'$ can be regarded as the new query or value. For an example of query $Q = X W_{q}$, the updated query $Q' = Q + s \cdot X W_{down}W_{up}$, which has a similar updating formulation with Adapter. To introduce the graph structure information ($S \in \mathbb{R}^{n \times n}$) in Sec.~\ref{subsec:can_graph}, we modify Eq.~(\ref{eq:lora}) as follows:
\begin{equation}
    X' = XW_{\ast} + s \cdot S X W_{down} W_{up}
    \label{eq:lora_modify}
\end{equation}

\paragraph{BitFit} \citet{ben-zaken2022bitfit} employ a straightforward strategy to expose knowledge of the pre-trained models for downstream tasks via tuning the bias terms ($b$) of the model. To be specific, for the linear operation, the updated output is equal to:
\begin{equation}
    X' = XW + b
    \label{eq:bitfit}
\end{equation}
where $W$ refers to the pre-trained weights of the linear layer, and $b \in \{b_{q}, b_{k}, b_{v}, b_{o}, b_{f1}, b_{f2}\}$ (including the parameters of LayerNorm layers). Note that only $b$ is updated during fine-tuning.

\paragraph{Hyperformer} This method can be regarded as a variant of Adapter via using shared hypernetworks in multi-task scenarios. Specifically, Hyperformer \cite{karimimahabadi2021parameterefficient} leverages the task conditioned hypernetworks to obtain the parameters of Adapter modules, i.e.:
\begin{equation}
    X' = X + \mathrm{LN}\big(\sigma(X W_{down}) W_{up}\big)
    \label{eq:hyperformer}
\end{equation}
where $W_{down} = W^{D} I_{\tau}, W_{up} = W^{U} I_{\tau}$, $W^{D} \in \mathbb{R}^{(d \times r) \times t}, W^{U} \in \mathbb{R}^{(r \times d) \times t}$, and $I_{\tau} \in \mathbb{R}^{t}$ is task embedding for each individual task ($\tau$). Similar operations are also conducted on the parameters of LayerNorm layer $\mathrm{LN}(\cdot)$.

\paragraph{Compacter} \citet{mahabadi2021compacter} introduce the Kronecker product and weights sharing tricks to reduce the ratio of trainable parameters in Adapter. Specifically, for the learnable weight $W \in \mathbb{R}^{d \times r}$, we can decompose $W$ into multiple ``small'' matrices via Kronecker product ($\otimes$):
\begin{equation}
    W = \sum_{i=1}^{n} A_{i} \otimes B_{i}
    \label{eq:compacter}
\end{equation}
where $A_{i} \in \mathbb{R}^{n \times n}, B_{i} \in \mathbb{R}^{\frac{d}{n} \times \frac{r}{n}}$. This decomposition method can be applied to the trainable weights $W_{down}$ and $W_{up}$ in Adapter. Besides, Compacter also shares the $A_{i}$ across all Adapter blocks.

\paragraph{MAM} \citet{he2022unified} investigate the traditional PEFTs from three perspectives: updated functional form, insertion form, and modified representation, and then offer a unified view to understand existing PEFTs: $X' = X + \Delta X$, where $\Delta X$ is learned by PEFT modules. Furthermore, based on their findings (e.g., FFN can better utilize modification than MHA at larger capacities), they propose a new PEFT method (MAM) by combining the most optimal choices.

\subsection{More Descriptions about Datasets}
\label{app:datasets}
We evaluate our proposed method and other baselines on nine benchmark datasets: MolHIV, MolPCBA, FreeSolv, ESOL, BBBP, Estrogen-$\alpha$, Estrogen-$\beta$, MetStab$_\mathrm{low}$, and MetStab$_\mathrm{high}$ \cite{gaulton2012chembl, hu2021open, podlewska2018metstabon, wu2018moleculenet}. Following the previous settings \cite{maziarka2020molecule, ying2021transformers}, we split the dataset into the training set, the validation set, and the test set in the ratio of $8: 1: 1$, and the statistical information is shown in Tab.~\ref{tab:sta_datasets}.
Specifically, for each of dataset, MolHIV and MolPCBA are two molecular property prediction datasets, which are derived from the popular graph benchmark OGB. The target of this task is to predict the binary labels for each molecule, which indicates whether it has a particular property or not.
FreeSolv and ESOL are regression tasks for predicting water solubility in terms of hydration free energy and log solubility.
BBBP is a binary classification task for predicting the ability of a molecule to penetrate the blood-brain barrier.
The aim of Estrogen-$\alpha$ and Estrogen-$\beta$ is to predict whether a compound is active towards a given target based on experimental data from the ChEMBL database.
Last, MetStab$_\mathrm{low}$ and MetStab$_\mathrm{high}$ are also binary classification tasks for predicting whether a compound has high or low metabolic stability.

\begin{table}[h]
    \centering
    \caption{Statistics for different datasets. ``\# Train'', ``\# Valid'', and ``\# Test'' indicate the number of training, validation, and test sets, respectively. Here, ``Clf.'' and ``Reg.'' refer to the classification and regression tasks, respectively.}
    \scalebox{0.77}{
        \begin{tabular}{lccccccccc}
            \toprule
                 Datasets & MolHIV & MolPCBA & FreeSolv & ESOL & BBBP & Estrogen-$\alpha$ & Estrogen-$\beta$ & MetStab$_\mathrm{low}$ & MetStab$_\mathrm{high}$ \\
            \midrule
                \# Train  & 32,901 & 350,343 & 513  & 902   & 1,631 & 1,918 & 1,568 & 1,701 & 1,701 \\
                \# Valid  & 4,113  & 43,793  & 64   & 113   & 204   & 240   & 196   & 213   & 213   \\
                \# Test   & 4,113  & 43,793  & 65   & 113   & 204   & 240   & 197   & 213   & 213   \\
                Task Type & Clf.   & Clf.    & Reg. & Reg.  & Clf.  & Clf.  & Clf.  & Clf.  & Clf.  \\
                Metric    & AUC    & AP      & RMSE & RMSE  & AUC   & AUC   & AUC   & AUC   & AUC   \\
            \bottomrule
        \end{tabular}
    }
    \label{tab:sta_datasets}
\end{table}

\begin{table}[h]
    \centering
    \caption{The detailed experimental configurations (batch size, learning rate, and bottleneck size) of various methods on a range of datasets, where Full-FT denotes full fine-tuning.}
    \scalebox{0.73}{
        \begin{tabular}{lccccccccc}
            \toprule
                 Method & MolHIV & MolPCBA & FreeSolv & ESOL & BBBP & Estrogen-$\alpha$ & Estrogen-$\beta$ & MetStab$_\mathrm{low}$ & MetStab$_\mathrm{high}$ \\
            \midrule
                \multicolumn{10}{c}{Batch Size / Learning Rate} \\
            \midrule
                Full-FT & 128 / 2e-5 & 128 / 2e-5 & 32 / 1e-5 & 32 / 1e-5 & 32 / 1e-5 & 32 / 1e-5 & 32 / 1e-5 & 32 / 1e-5 & 32 / 1e-5 \\
                Adapter & 128 / 2e-3 & 128 / 2e-3 & 64 / 1e-3 & 64 / 2e-3 & 64 / 2e-3 & 32 / 1e-3 & 64 / 1e-3 & 32 / 2e-3 & 64 / 2e-3 \\
            Hyperformer & 128 / 2e-3 & 128 / 1e-3 & 64 / 2e-3 & 32 / 1e-3 & 32 / 2e-3 & 32 / 1e-3 & 32 / 2e-3 & 64 / 1e-3 & 32 / 1e-3 \\
            Compacter   & 128 / 1e-3 & 128 / 2e-3 & 32 / 1e-3 & 32 / 1e-3 & 32 / 2e-3 & 64 / 2e-3 & 32 / 2e-3 & 64 / 2e-3 & 32 / 2e-3 \\
                MAM     & 128 / 2e-3 & 128 / 1e-3 & 32 / 1e-3 & 64 / 1e-3 & 32 / 1e-3 & 64 / 2e-3 & 32 / 1e-3 & 32 / 2e-3 & 64 / 2e-3 \\
                LoRA    & 128 / 1e-3 & 128 / 2e-3 & 32 / 2e-3 & 32 / 1e-3 & 64 / 2e-3 & 64 / 2e-3 & 32 / 1e-3 & 32 / 1e-3 & 32 / 1e-3 \\
                BitFit  & 128 / 1e-3 & 128 / 1e-3 & 32 / 1e-3 & 32 / 1e-3 & 32 / 1e-3 & 32 / 1e-3 & 32 / 1e-3 & 32 / 1e-3 & 32 / 1e-3 \\
              G-Adapter & 128 / 2e-3 & 128 / 1e-3 & 32 / 1e-3 & 64 / 2e-3 & 64 / 1e-3 & 32 / 2e-3 & 32 / 2e-3 & 32 / 2e-3 & 32 / 2e-3 \\
            \midrule
                \multicolumn{10}{c}{Bottleneck Size ($r$)} \\
            \midrule
                Adapter & 16 & 64 & 16 & 8 & 16 & 4 & 32 & 128 & 64 \\
            Hyperformer & 8  & 48 & 16 & 4 & 16 & 8 & 16 & 64  & 48 \\
            Compacter   & 16 & 48 & 32 & 8 & 16 & 4 & 16 & 48  & 32 \\
                MAM     & 8  & 32 & 8 & 16 & 32 & 4 & 8  & 32  & 32 \\
                LoRA    & 4  & 32 & 16 & 8 & 32 & 8 & 4  & 4   & 16 \\
              G-Adapter & 4  & 48 & 4 & 16 & 4  & 4 & 4  & 16  & 64 \\
            \bottomrule
        \end{tabular}
    }
    \label{tab:configs}
\end{table}

\subsection{Detailed Experimental Configurations and Implementations}
\label{app:config}
For a variety of methods, we perform a relatively fair hyper-parameter search in terms of learning rate and batch size from $\{1e-3, 2e-3, 1e-4, 1e-4, 1e-5, 2e-5\}$ and $\{32, 64, 128\}$, respectively. For the PEFTs with bottleneck architecture, we select the bottleneck size ($r$) from $\{4, 8, 16, 32, 48, 64, 128\}$ on various datasets. The detailed configurations are reported in Tab.~\ref{tab:configs}. In addition, we also provide some general pseudo-code to illustrate how to integrate our proposed method into existing GTNs in Alg.~\ref{alg:pseudo_code}.

\subsection{More Experimental Results}
\label{app:exp}
\begin{figure}[h]
    \centering
    \includegraphics[width=0.6\textwidth]{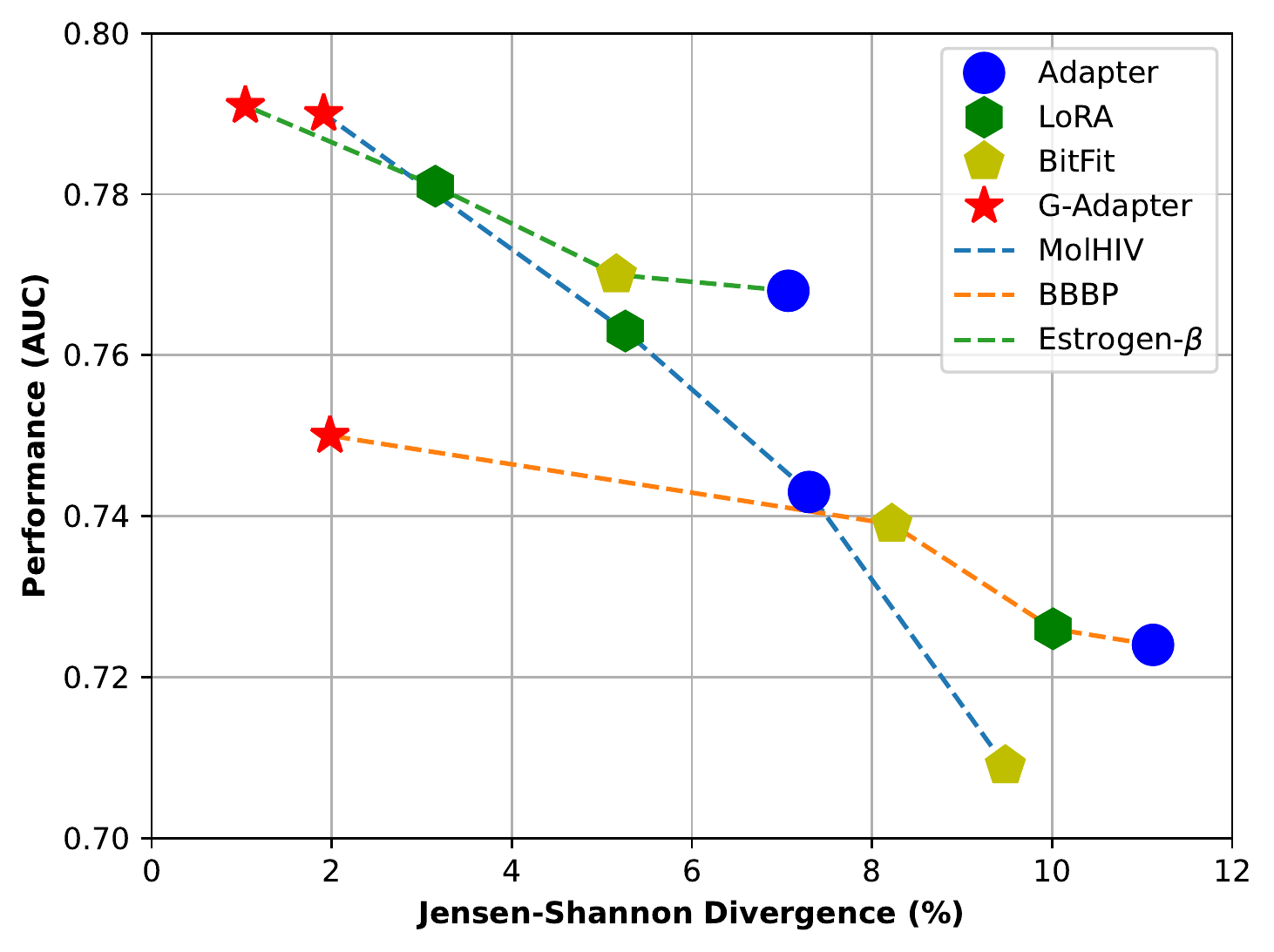}
    \caption{The relationship between Jensen-Shannon divergence and performance.}
    \label{fig:js_vs_perf}
\end{figure}

\begin{figure}[h]
    \centering
    \subfigure[FreeSolv]{
        \includegraphics[width=0.3\textwidth]{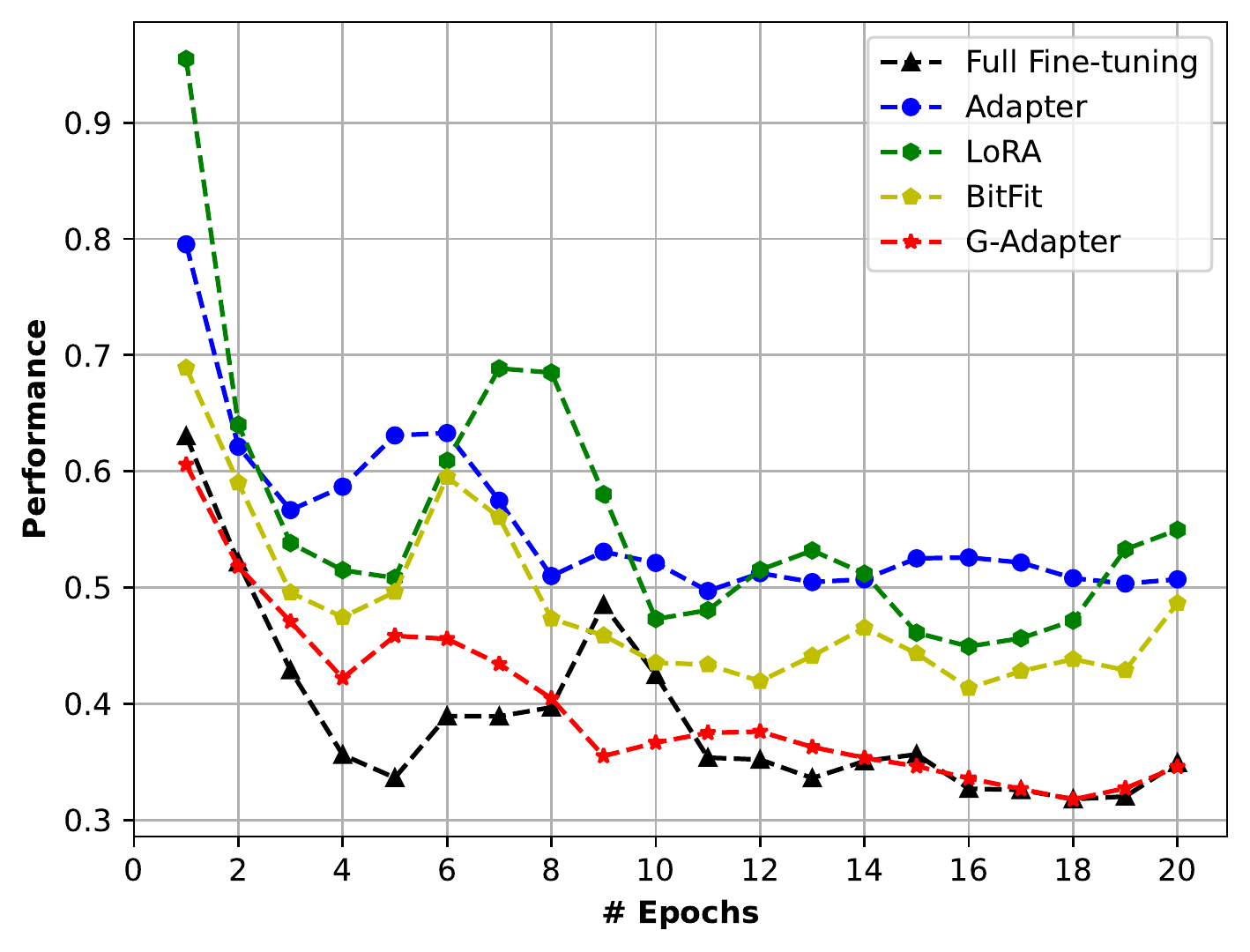}
    }
    \subfigure[ESOL]{
        \includegraphics[width=0.3\textwidth]{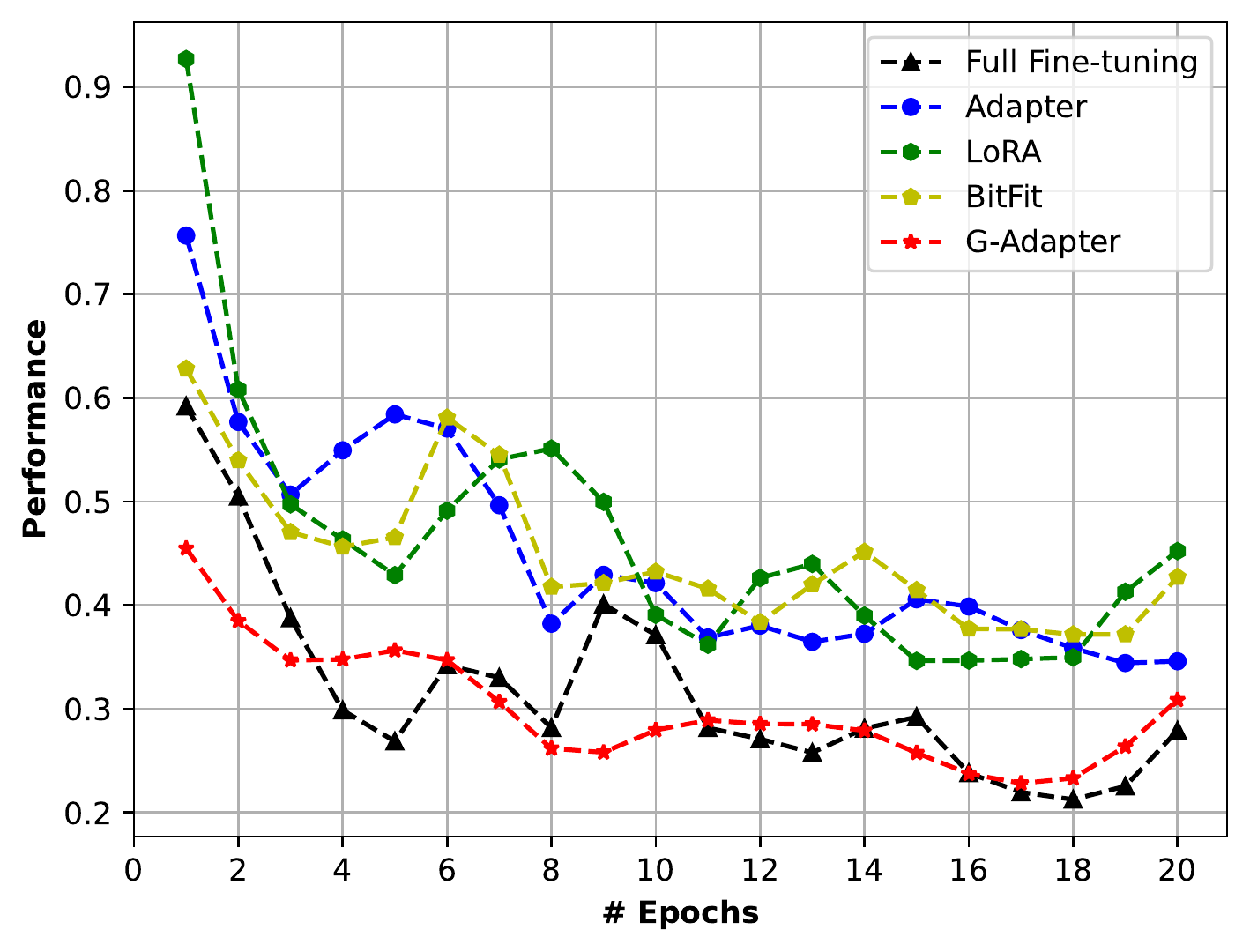}
    }
    \subfigure[BBBP]{
        \includegraphics[width=0.3\textwidth]{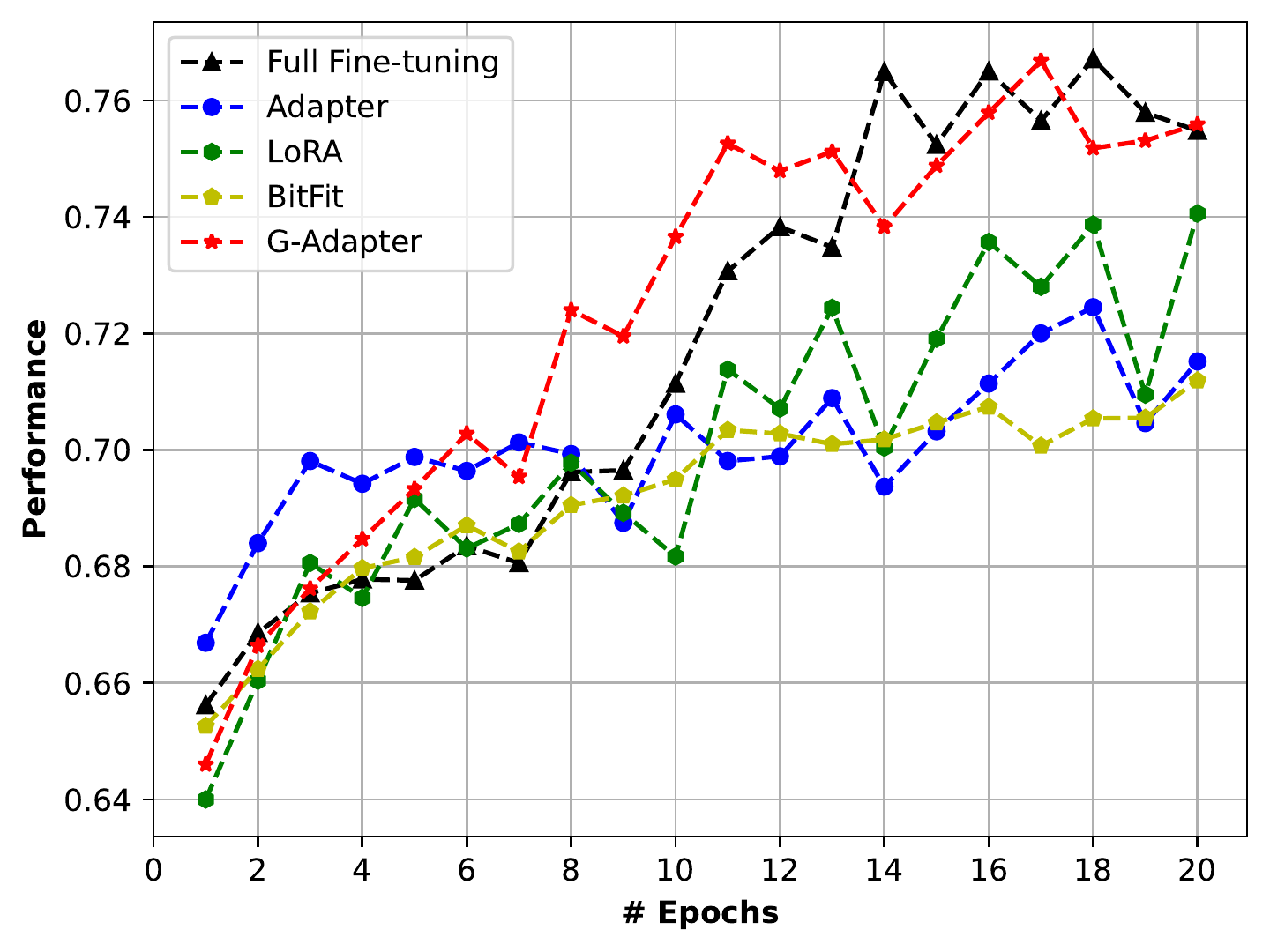}
    } \\
    \subfigure[Estrogen-$\beta$]{
        \includegraphics[width=0.3\textwidth]{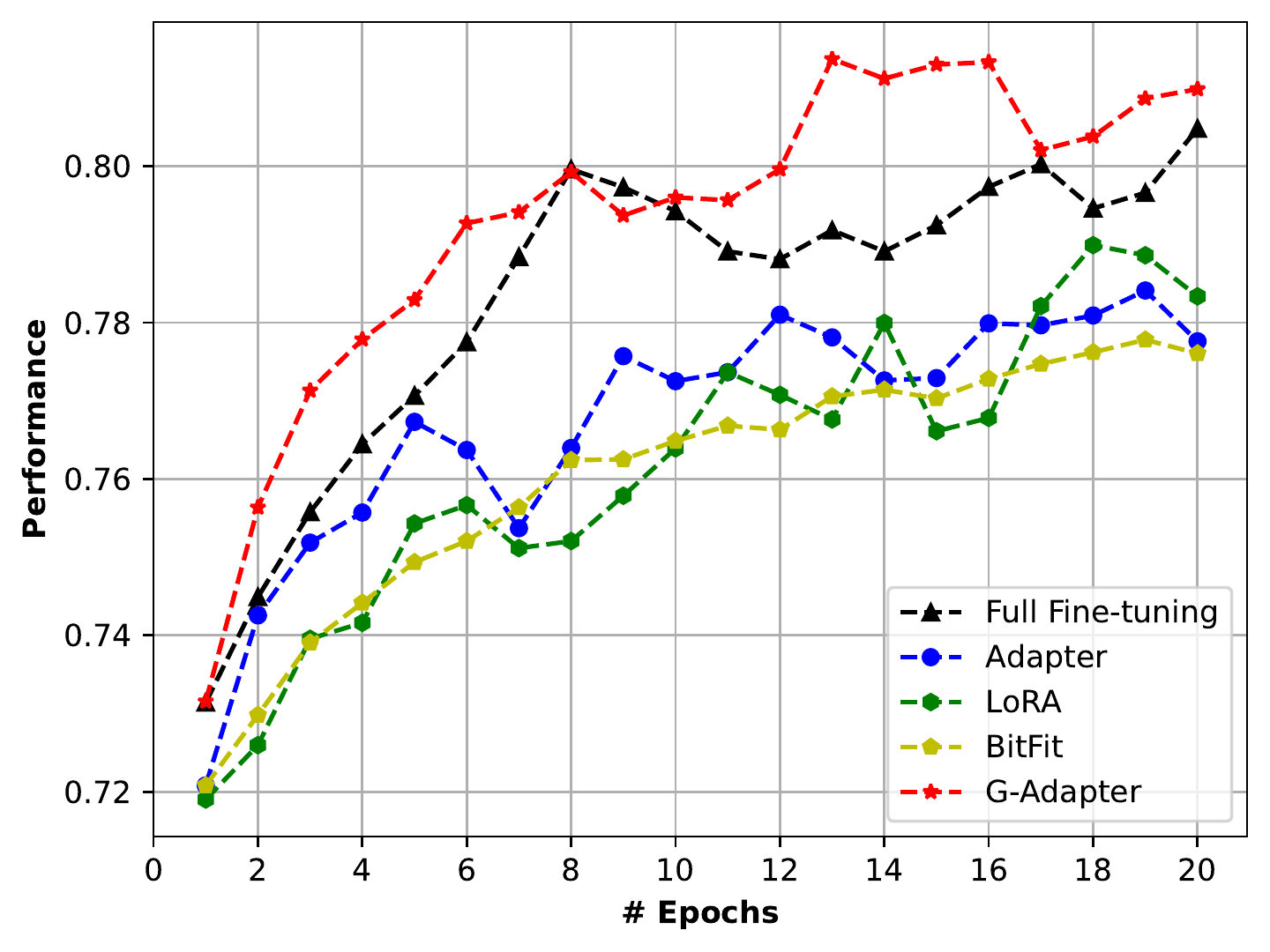}
    }
    \subfigure[MetStab$_\mathrm{low}$]{
        \includegraphics[width=0.3\textwidth]{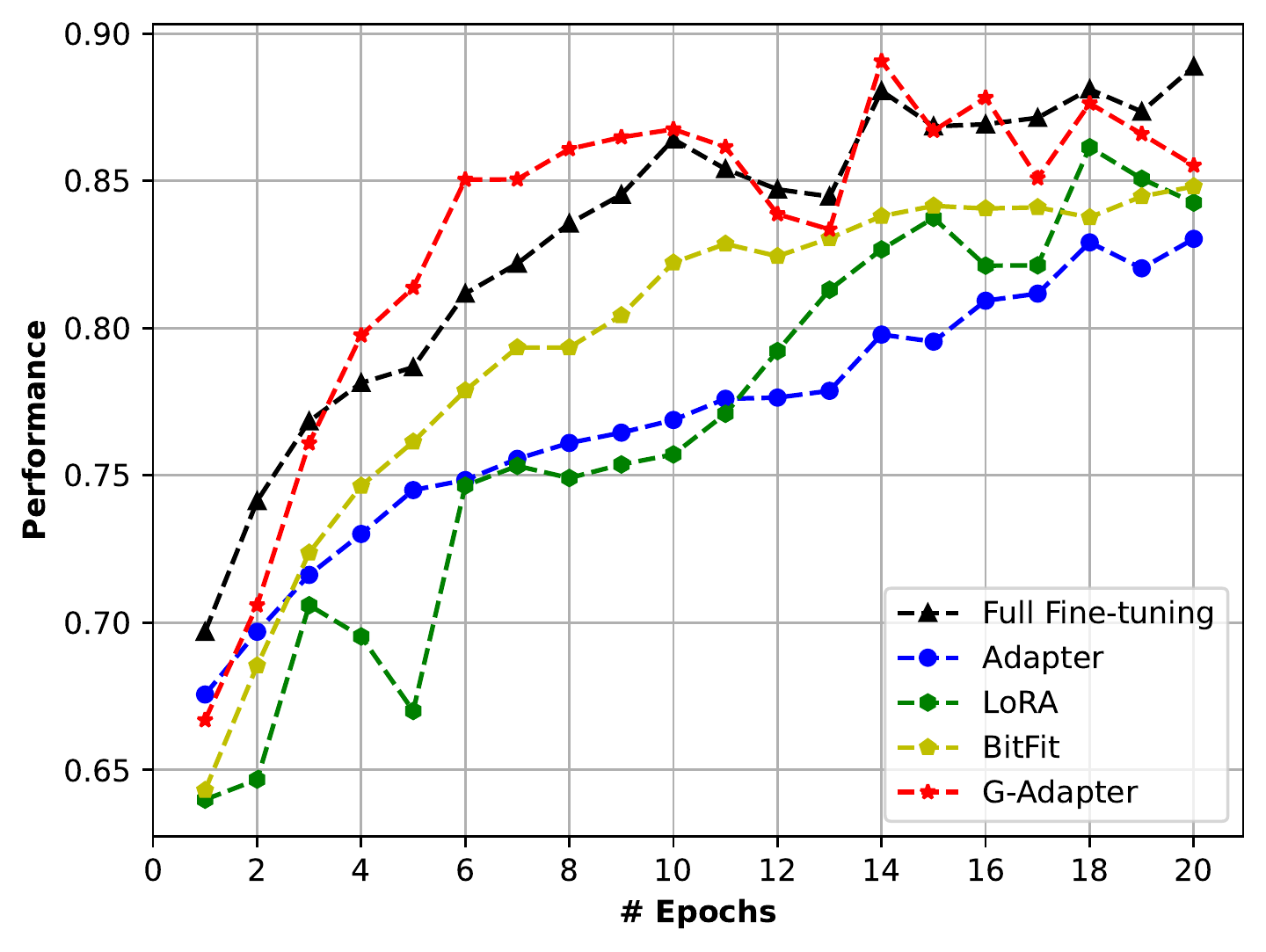}
    }
    \subfigure[MetStab$_\mathrm{high}$]{
        \includegraphics[width=0.3\textwidth]{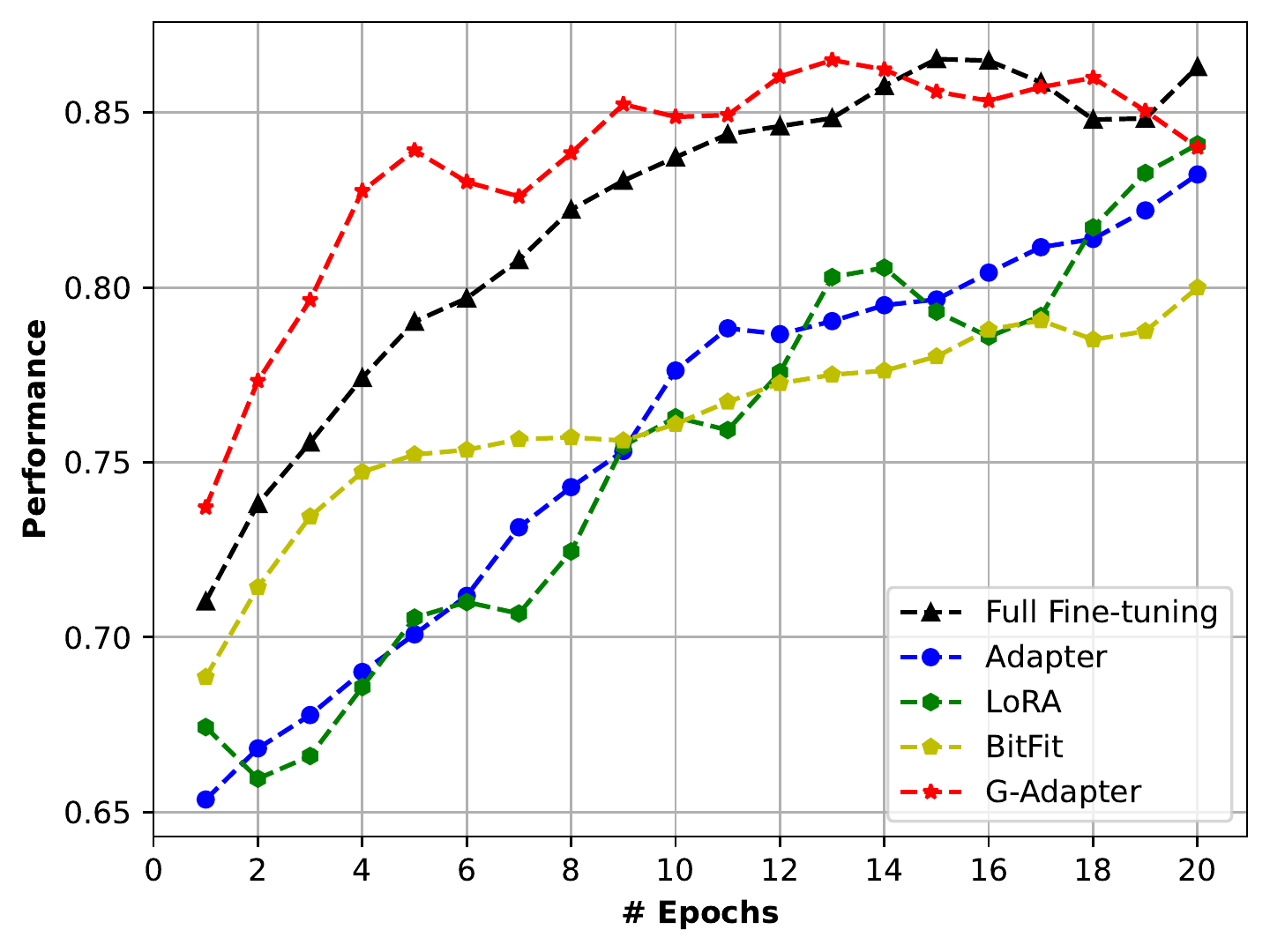}
    }
    \caption{More comparisons of training efficiency between PEFTs and full fine-tuning on more datasets. Note that the evaluation protocol for FreeSolv and ESOL is RMSE (the lower, the better), while the others are AUC (the higher, the better).}
    \label{fig:more_compare_train_eff}
\end{figure}

\begin{figure}[h]
    \centering
    \includegraphics[width=1\textwidth]{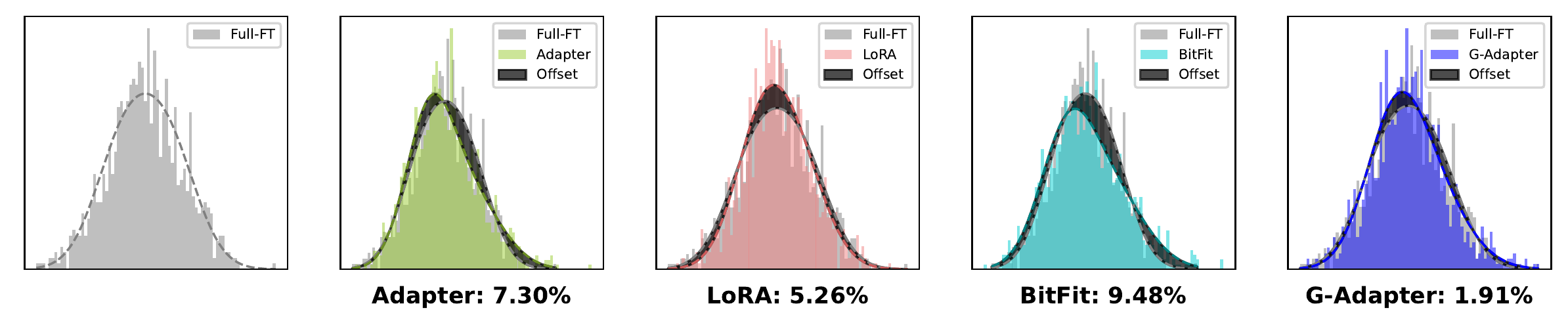}
    \caption{Illustration of feature distribution shift on MolHIV with pre-trained Graphormer.}
    \label{fig:feat_dist_shift_ext2}
\end{figure}

\begin{figure}[h]
    \centering
    \includegraphics[width=1\textwidth]{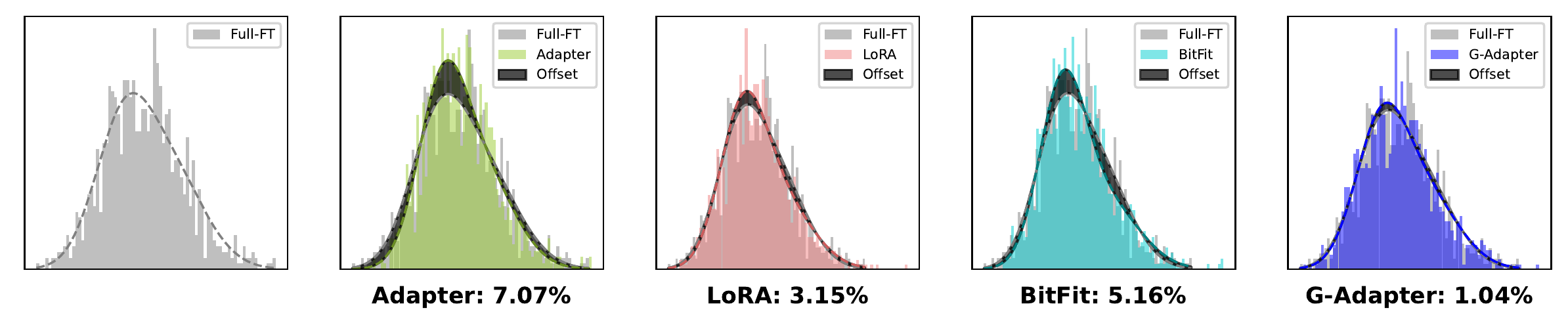}
    \caption{Illustration of feature distribution shift on Estrogen-$\beta$ with pre-trained MAT.}
    \label{fig:feat_dist_shift_ext1}
\end{figure}

\begin{figure}[h]
    \centering
    \includegraphics[width=1\textwidth]{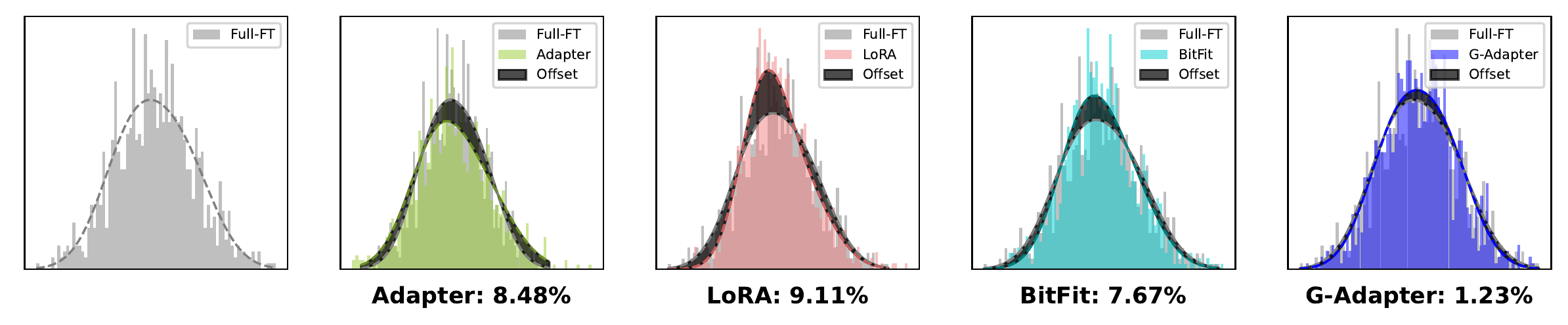}
    \caption{Illustration of feature distribution shift on MetStab$_\mathrm{low}$ with pre-trained MAT.}
    \label{fig:feat_dist_shift_ext3}
\end{figure}

Here, we conduct more experiments to demonstrate the feature distribution shift issue on different datasets (MolHIV, Estrogen-$\beta$, and MetStab$_\mathrm{low}$) with two pre-trained GTNs (Graphormer and MAT). The experimental results are shown in Fig.~\ref{fig:feat_dist_shift_ext2}, \ref{fig:feat_dist_shift_ext1}, and \ref{fig:feat_dist_shift_ext3}. In addition, we depict the relationship between Jensen-Shannon divergence (which measures the degree of discrepancy in feature distribution between PEFT and full fine-tuning) and performance across various datasets and methods in Fig.~\ref{fig:js_vs_perf}. 
We also supplement more comparisons in terms of the training efficiency, the impact of different designs, and the effect of graph structure information for traditional PEFTs on more datasets. The results are shown in Fig.~\ref{fig:more_compare_train_eff}, Tab.~\ref{tab:more_compare_can_graph}, \ref{tab:more_compare_diff_des}, where we could draw the consistent conclusion as before.

\begin{table}[h]
    \centering
    \caption{The impact of graph structure information on traditional PEFTs.}
    \scalebox{0.75}{
        \begin{tabular}{lccccc}
            \toprule
                 Method & FreeSolv & ESOL & Estrogen-$\alpha$ & Estrogen-$\beta$ & MetStab$_\mathrm{high}$ \\
            \midrule
                Adapter       & 0.327 & 0.320 & 0.978 & 0.768 & 0.859 \\
                Adapter + $S$ & 0.311 & 0.314 & 0.980 & 0.773 & 0.866 \\
            \midrule
                LoRA       & 0.309 & 0.284 & 0.979 & 0.781 & 0.878 \\
                LoRA + $S$ & 0.297 & 0.280 & 0.978 & 0.789 & 0.880 \\
            \bottomrule
        \end{tabular}
    }
    \label{tab:more_compare_can_graph}
\end{table}

\begin{table}[h]
    \centering
    \caption{The impact of insertion position and components on performance.} 
    \scalebox{0.85}{
        \begin{tabular}{lccccccc}
            \toprule
                 Method & FreeSolv & ESOL & BBBP & Estrogen-$\alpha$ & Estrogen-$\beta$ & MetStab$_\mathrm{low}$ & MetStab$_\mathrm{high}$ \\
            \midrule
                G-Adapter                 & 0.280 & 0.279 & 0.750 & 0.976 & 0.791 & 0.865 & 0.881 \\
            \midrule
                G-Adapter (pre\_mha)      & 0.317 & 0.312 & 0.739 & 0.966 & 0.772 & 0.832 & 0.865 \\
                G-Adapter (post\_mha)     & 0.304 & 0.298 & 0.733 & 0.963 & 0.787 & 0.847 & 0.863 \\
                G-Adapter (pre\_ffn)      & 0.291 & 0.281 & 0.741 & 0.973 & 0.789 & 0.873 & 0.875 \\
                G-Adapter (post\_ffn)     & 0.289 & 0.284 & 0.739 & 0.974 & 0.788 & 0.869 & 0.872 \\
                G-Adapter (mha + ffn)     & 0.294 & 0.296 & 0.735 & 0.972 & 0.777 & 0.864 & 0.869 \\
            \midrule
                G-Adapter (w/o. $S$)      & 0.351 & 0.335 & 0.706 & 0.950 & 0.737 & 0.823 & 0.832 \\
                G-Adapter (w/o. pre\_ln)  & 0.321 & 0.319 & 0.711 & 0.962 & 0.745 & 0.852 & 0.841 \\
                G-Adapter (w/o. post\_ln) & 0.314 & 0.323 & 0.713 & 0.955 & 0.751 & 0.846 & 0.845 \\
                G-Adapter (w/o. ln)       & 0.343 & 0.328 & 0.705 & 0.956 & 0.726 & 0.833 & 0.839 \\
                G-Adapter (w/o. act\_fn)  & 0.331 & 0.301 & 0.713 & 0.962 & 0.755 & 0.844 & 0.855 \\
                G-Adapter (w/o. breg)     & 0.346 & 0.325 & 0.704 & 0.951 & 0.743 & 0.849 & 0.834 \\
            \bottomrule
        \end{tabular}
    }
    \label{tab:more_compare_diff_des}
\end{table}

\begin{algorithm}
\caption{Pseudo-code of G-Adapter in a PyTorch-like style.}
\begin{lstlisting}
# Define the G-Adapter block
class GAdapter(nn.Module):
    def __init__(self, hidden_size, bottleneck_size)
        super(GAdapter, self).__init__()
        self.down = nn.Linear(hidden_size, bottleneck_size)
        self.up   = nn.Linear(bottleneck_size, hidden_size)
        self.pre_ln  = nn.LayerNorm(hidden_size)
        self.post_ln = nn.LayerNorm(hidden_size)
        self.act_fn  = nn.ReLU()

    def forward(self, x, s):
        # x: batch_size * sequence_length * hidden_size
        # s: batch_size * sequence_length * sequence_length
        x = self.pre_ln(x)
        x = self.act_fn(self.up(self.down(torch.matmul(s, x)))) + x
        x = self.post_ln(x)
        return x

# Apply the G-Adapter block into the built-in module
class Encoder(nn.Module):
    def __init__(self, hidden_size, bottleneck_size, *args, **kwargs)
        super(Encoder, self).__init__()
        ...
        # +++ #
        self.gadapter = GAdapter(hidden_size, bottleneck_size)
        # +++ #
        ...

    def forward(self, x, s):
        ...
        # +++ #
        x = self.gadapter(x, s)
        # +++ #
        ...
        return x
\end{lstlisting}
\label{alg:pseudo_code}
\end{algorithm}

\end{document}